\PassOptionsToPackage{table}{xcolor}
\documentclass[journal]{IEEEtran}
\usepackage[utf8]{inputenc}

\usepackage[hidelinks]{hyperref}

\usepackage{amsmath,amsfonts}
\usepackage{algorithm}
\usepackage{array}
\usepackage[caption=false,font=normalsize,labelfont=sf,textfont=sf]{subfig}
\usepackage{textcomp}
\usepackage{stfloats}
\usepackage{url}
\usepackage{verbatim}
\usepackage{graphicx}
\usepackage{cite}
\usepackage{orcidlink}
\usepackage{glossaries}
\DeclareUnicodeCharacter{2212}{-}
\usepackage{multirow}
\usepackage[table]{xcolor}
\usepackage{tabularx}
\usepackage{booktabs}
\usepackage{multirow}

\definecolor{lightcoral}{RGB}{240,128,128}
\definecolor{lightgreen}{RGB}{144,238,144}

\newacronym{nhtsa}{NHTSA}{National Highway Traffic Safety Administration}
\newacronym{ecg}{ECG}{Electrocardiogram}
\newacronym{dqn}{DQN}{Deep Q-Network}
\newacronym{lstm}{LSTM}{Long Short-Term Memory}
\newacronym{cnn}{CNN}{Convolutional Neural Network}
\newacronym{cnn-lstm}{CNN-LSTM}{Convolutional Neural Network - Long Short-Term Memory }
\newacronym{ae-lstm}{AE-LSTM}{Autoencoder - Long Short-Term Memory}
\newacronym{mdp}{MDP}{Markov Decision Process}
\newacronym{hrv}{HRV}{Heart Rate Variability}
\newacronym{rem}{REM}{Rapid Eye Movement}
\newacronym{eog}{EOG}{Electrooculography}
\newacronym{emg}{EMG}{Electromyography}
\newacronym{rl}{RL}{Reinforcement Learning}
\newacronym{qlearning}{Q-learning}{A Reinforcement Learning Algorithm}

\newacronym{carla}{CARLA}{Car Learning to Act}
\newacronym{dddqn}{DD-DQN}{Double Dual Deep Q-Network}
\newacronym{drl}{DRL}{Deep Reinforcement Learning}

\newacronym{eeg}{EEG}{Electroencephalography}
\newacronym{rri}{RRi}{RR Interval}
\newacronym{lf}{LF}{Low Frequency}
\newacronym{hf}{HF}{High Frequency}
\newacronym{lfhf}{LF/HF}{Low-Frequency to High-Frequency Ratio}
\newacronym{sdnn}{SDNN}{Standard Deviation of NN Intervals}
\newacronym{rmssd}{RMSSD}{Root Mean Square of Successive Differences}
\newacronym{pnn50}{pNN50}{Percentage of Successive RR Intervals Differing by More than 50 ms}
\newacronym{nk2}{NK2}{NeuroKit-2}
\newacronym{l}{L}{Window Size}
\newacronym{m}{M}{Overlap Percentage}
\newacronym{n}{N}{Number of Capsules}
\newacronym{c}{C}{Capsule Size}

\newacronym{ae}{AE}{Auto-Encoder}
\newacronym{nn}{NN}{Neural Network}
\newacronym{nns}{NNs}{Neural Networks}
\newacronym{duelingdqn}{Dueling DQN}{Dueling Deep Q-Network}
\newacronym{per}{PER}{Prioritized Experience Replay}

\newacronym{lidar}{LiDAR}{Light Detection and Ranging}

\newacronym{api}{API}{Application Programming Interface}
\newacronym{dew}{DEW}{Drowsy Event Window}

\newacronym{rnn}{RNN}{Recurrent Neural Network}

\newacronym{cnm}{CNM}{CNM}

\newacronym{dddb}{DD-DB}{Drivers Drowsiness Database}

\newacronym{edf}{EDF}{European Data Format}

\newacronym{dbscan}{DBSCAN}{Density-Based Spatial Clustering of Applications with Noise}

\newacronym{gru}{GRU}{Gated Recurrent Unit}
\newacronym{pca}{PCA}{Principal Component Analysis}

\newacronym{ddd-qn}{DDD-QN}{Double Dual Deep Q-Network}
\newacronym{multi-config}{Multi-Config}{Multi-Configuration Approach}

\newacronym{speed}{v_t}{Speed of the vehicle at time t}
\newacronym{distance}{d_t}{Distance to the leading vehicle at time t}
\newacronym{relvelocity}{v_{\text{rel},t}}{Relative velocity between the ego vehicle and the leading vehicle}
\newacronym{drowsiness}{\theta_t}{Drowsiness level of the driver at time t}
\newacronym{acceleration}{Acceleration}{Action that increases vehicle speed}
\newacronym{neutral}{Neutral}{No control input for braking or acceleration}
\newacronym{braking}{Braking}{Action that decreases vehicle speed}
\newacronym{steering}{Steering}{Control of the vehicle's direction}
\newacronym{tireroadinteraction}{Tire-Road Interaction}{Frictional interaction between the vehicle tires and the road surface}
\newacronym{friction}{Friction}{Force opposing relative motion between the tire and the road surface}
\newacronym{rollingresistance}{Rolling Resistance}{Resistance encountered by the tire rolling on a surface}
\newacronym{suspension}{Suspension}{System of springs, shock absorbers, and linkages connecting a vehicle to its wheels}
\newacronym{lateralforces}{Lateral Forces}{Forces acting on the vehicle perpendicular to its direction of motion}
\newacronym{collisionhandling}{Collision Handling}{Mechanism to detect and handle collisions in the simulation}
\newacronym{carlasimulator}{CARLA Simulator}{CARLA, an open-source autonomous driving simulator}
\newacronym{actiondelay}{\Delta t}{Time delay applied to the agent's action due to drowsiness}
\newacronym{reactiontime}{d_{\text{reaction}}}{Reaction time delay associated with driver drowsiness}
\newacronym{collision}{Collision}{An event where the vehicle impacts an object}
\newacronym{pressurechange}{Abrupt Pressure Change}{Sudden change in braking pressure}
\newacronym{unsafedistance}{Unsafe Distance}{Distance to the leading vehicle that is too short to avoid collision}
\newacronym{safedistance}{Safe Distance}{Distance to the leading vehicle that is safe to maintain}
\newacronym{smoothbraking}{Smooth Braking}{Gradual application of braking force to avoid abrupt stops}
\newacronym{dodqn}{Double-DQN}{Double Deep Q-Network}
\newacronym{dudqn}{Dual-DQN}{Dual Deep Q-Network}
\newacronym{cv}{CV}{Cross-validation}
\newacronym{noisy dqn}{Noisy DQN}{Noisy Deep Q-Network}
\newacronym{ttc}{TTC}{Time-to-collision}
\newacronym{ddpg}{DDPG}{Deep Deterministic Policy Gradient}
\newacronym{pb}{PB}{principle-based}
\newacronym{td}{TD}{Temporal-Difference}
\newacronym{abs}{ABS}{Anti-Lock Braking System}
\newacronym{nsrw}{NSRW}{Normal Sinus Rhythm Window}
\newacronym{fifo}{FIFO}{First-In-First-Out}
\newacronym{rfe}{RFE}{Recursive Feature Elimination}
\newacronym{mi}{MI}{Mutual Information}
\newacronym{rf}{RF}{Random Forest Feature Importance}
\newacronym{pi}{PI}{Permutation Importance}
\newacronym{fdqn}{FDQN}{Fixed Q-Targets}

\newacronym{ans}{ANS}{Autonomic Nervous System}
\newacronym{tde}{TDE}{Time Delay Embedding}
\newacronym{v2i}{V2I}{Vehicle-to-Infrastructure}
\newacronym{dbif}{DBIF}{Drowsiness Braking Impact Factor}
\newacronym{can}{CAN}{Controller Area Network}

\begin{document}

\title{Drowsiness-Aware Adaptive Autonomous Braking System based on Deep Reinforcement Learning for Enhanced Road Safety}

\author{Hossem Eddine Hafidi \orcidlink{0009-0000-0520-4115}, 
        Elisabetta De Giovanni \orcidlink{0000-0003-3032-5140}, 
        Teodoro Montanaro \orcidlink{0000-0003-1750-8268}, 
        Ilaria Sergi \orcidlink{0000-0002-3797-3039}, \\
        Massimo De Vittorio \orcidlink{0000-0003-1601-6392}, 
        and Luigi Patrono \orcidlink{0000-0002-8591-1190}%

\thanks{Hossem Eddine Hafidi, Teodoro Montanaro, Ilaria Sergi, and Luigi Patrono are with the University of Salento, Lecce, Italy. (Emails: hossemed dine.hafidi@unisalento.it, hafhousedd@hotmail.com, teodoro.montanaro@u nisalento.it, ilaria.sergi@unisalento.it, luigi.patrono@unisalento.it.)}%
\thanks{Massimo De Vittorio and Hossem Eddine Hafidi are with Istituto Italiano di Tecnologia (IIT), Lecce, Italy. (Emails: massimo.devittorio@iit.it, hossem.hafidi@iit.it.)}%
\thanks{Elisabetta De Giovanni is with the Basque Center for Applied Mathematics (BCAM), Bilbao, Spain. (Email: edegiovanni@bcamath.org)}%
\thanks{Digital Object Identifier~00.0000/XYZ.0000.0000000}

}

\maketitle



\begin{abstract}
Driver drowsiness significantly impairs the ability to accurately judge safe braking distances. According to the European Transport Safety Council, fatigue is a contributing factor in approximately 10\% to 20\% of road accidents across Europe. Traditional driver-assistance systems generally lack adaptability to real-time physiological states such as drowsiness. In this paper, we address this challenge by developing a deep reinforcement learning-based autonomous braking system that integrates vehicle and driver data. The system detects drowsiness using physiological data and adapts the braking behavior in response to real-time vehicle dynamics and traffic conditions. The entire framework is implemented and evaluated within a high-fidelity simulation environment. The reinforcement learning agent leverages the latest advancements in Deep Q-Network (DQN) technologies, particularly Double and Dual networks. To incorporate drowsiness, we apply exhaustive benchmark analysis on 2-minute drowsiness windows extracted from ECG signals, considering different configurations of window segmentation and overlapping to find the most effective setup. Moreover, we select a Recurrent Neural Network (RNN) architecture as the optimal model for our case scenario. The drowsiness output detected in real-time by our RNN model is used as input to the observable state space of the DQN agent, where the driver error during drowsiness is simulated by an action delay. The DQN agent with double-dual networks achieves a 99.99\% success rate in adjusting brakes and evading accidents across both drowsy and non-drowsy scenarios. Our approach to integrating vehicle and driver data can contribute to building physiology-enhanced adaptive intelligent driving systems to enhance safety on the road.
\end{abstract}

\begin{IEEEkeywords}
Automotive safety, Driver physiology, Deep reinforcement learning, Vehicle braking systems.
\end{IEEEkeywords}

\section{Introduction}

\IEEEPARstart{D}{riving} demands continuous attention and rapid decision-making, particularly when executing safety-critical maneuvers such as braking. The effectiveness of these maneuvers relies on the driver's awareness, the vehicle’s dynamic capabilities, and the surrounding traffic context. One key factor affecting awareness is drowsiness, which is a transitional state between wakefulness and sleep. Drowsiness impairs cognitive performance and ultimately delays reaction times~\cite{2013Drowsy,WU2023107590}.  According to the European Commission, drowsiness contributes to 15--20\% of serious road accidents, with drowsy drivers experiencing a 29\% higher crash risk than alert drivers~\cite{Fredriksson2021}. Although traditional braking systems such as Anti-lock Braking Systems (ABS), Electronic Stability Control (ESC), and Adaptive Cruise Control (ACC) have substantially improved vehicle safety through dynamic modeling, they remain limited in their ability to adapt to real-time driver cognitive states during vehicle control~\cite{9864310}. To the best of our knowledge, current autonomous systems lack integration of physiological monitoring for adapting real-time braking under drowsy conditions.

Nevertheless, the identification of drowsiness during driving has been extensively explored. Drowsiness detection in the automotive domain utilizes various methods, including visual sleep scoring and physiological signal analysis \cite{Chowdhury2018Sensor}. Among these signals, \gls{ecg} is particularly effective for identifying 'awake' and 'asleep' states due to its influence from the autonomous nervous system, and its non-invasive nature. It provides key physiological features such as \gls{hrv}, whose temporal and sequential patterns are shaped by the signal segmentation process \cite{WU2023107590}. Optimizing temporal segmentation through sliding windows is therefore essential for accurate drowsiness detection. However, most existing approaches only focus on detection and do not directly influence braking actions. Integrating drowsiness detection into vehicle control is essential to mitigate impaired braking caused by delayed reaction times.

In this work, we address these challenges by proposing a physiology-aware autonomous braking system designed for real-world deployment, integrating real-time drowsiness detection into a \gls{drl} framework. Specifically, a \gls{dddqn} agent is trained to learn adaptive braking policies that account for vehicle dynamics, traffic conditions, and driver cognitive state. Training and evaluation take place in an environment designed to mirror real-world driving conditions. Our contributions include an optimized window-shifting method for \gls{ecg} segmentation, where multiple temporal configurations are evaluated to identify the most effective strategy. The driver’s drowsiness state is inferred using a \gls{rnn} trained on \gls{ecg}-derived features. The learned drowsiness predictions are incorporated into the agent’s observable state space, where the detected drowsiness status is designed to delay the execution of control actions to emulate an impaired driver. To the best of our knowledge, this is the first system to integrate \gls{ecg}-based drowsiness detection with \gls{drl} braking and delayed action modeling, designed for scenarios representative of real-world cases. We benchmark the \gls{dddqn} agent against three baseline architectures: vanilla \gls{dqn}, \gls{dodqn}, and \gls{dudqn}. Our proposed agent achieves a 99.99\% success rate in maintaining safe following distances and avoiding collisions under both alert and drowsy conditions. Across 30,000 seconds of testing, only 0.9 seconds of cumulative violations of safe distance were recorded, all occurring under drowsy states, which demonstrates the agent’s robustness and ability to adapt to impaired conditions.

The rest of this paper is structured as follows. Section~\ref{sec:Background and Related Work} reviews related works on drowsiness detection, \gls{hrv} analysis, and \gls{drl} for braking strategies. Section~\ref{sec:Proposed System} presents the proposed drowsiness-aware braking system, describing its architecture and the problem formalization. Section~\ref{sec:Feature extraction} details the preprocessing and classification of \gls{ecg} signals for drowsiness detection, along with the training methodology of the \gls{rnn}-based model. Section~\ref{sec:DQN based Deep RL Framework} explains the \gls{dddqn} training framework, including the implementation of drowsiness-induced action delays. Section~\ref{sec:Experiments and Results} reports the experimental setup and results evaluating the agent’s adaptive braking behavior. Finally, Section~\ref{sec:Discussion} discusses the findings and limitations, followed by concluding remarks.

\section{Background and Related Works}
\label{sec:Background and Related Work}

This section reviews the literature on intelligent braking systems and drowsiness detection, focusing on two main domains: quantified metrics of Drowsiness (Section~\ref{subsec:quantified_metrics_of_drowsiness}) and \gls{drl} for braking strategies (Section~\ref{subsec:deep_q_networks_and_braking_strategies}). While substantial research exists in both areas independently, to the best of our knowledge, no prior work has directly integrated real-time physiological monitoring with reinforcement learning-based braking strategies.

\subsection{Quantified Metrics of Drowsiness}
\label{subsec:quantified_metrics_of_drowsiness}

Drowsiness in drivers results from alterations in the \gls{ans}, which in turn affects multiple physiological systems, including respiration, cardiovascular activity, skin conductance, and visual perception. These changes produce observable physiological signals and signs such as respiratory rate, heart rate, and pupil behavior, which can be monitored to assess the driver’s cognitive state \cite{WU2023107590}. Traditional methods for detecting driver drowsiness involve visual analysis of sleep stages, often using optical or infrared cameras to monitor indicators like eye closure \cite{Schulz2008}. A key metric is the Percentage of Eyelid Closure over the Pupil over Time (PERCLOS), which increases during drowsy conditions \cite{zpad006}. However, visual methods can be obstructed by factors like glasses or poor lighting \cite{Ramzan2019A}. To address this, alternative hardware solutions such as smart glasses have been proposed to reduce reliance on unobstructed visual cues \cite{8493318}. Additionally, metrics like the Percentage of Yawning (POY) have been used when eye detection is unreliable \cite{Wang2024}.

Physiological analyses through \Gls{eeg} can capture brain wave patterns associated with drowsiness, particularly in Theta, Alpha, and Beta bands \cite{Strijkstra2003Subjective}. In \gls{eeg}-based monitoring systems, neural activity is typically acquired through a multi-electrode cap positioned on the scalp. Such configurations are prone to motion-induced artifacts, as driver movements can cause electrode displacement, thereby compromising signal integrity \cite{Casson2010Wearable}. In contrast, \gls{ecg}-based analysis offers a less intrusive alternative, as cardiac signals can be acquired using commercially available chest bands, without the need for specialized headgear. Moreover, the high amplitude of QRS complexes contributes to greater robustness against noise \cite{Fujiwara20242956}. 

\gls{hrv} refers to the variation in time intervals between consecutive heartbeats, and it is commonly used as a physiological marker of autonomic nervous system activity. The use of \Gls{hrv} features calculated from \gls{ecg} for drowsiness detection offers a practical approach, particularly when \gls{ecg} is acquired using a chest-mounted sensor during seated driving. In this context, the relatively limited movement of the chest reduces the likelihood of motion artifacts compared to signals like \gls{eeg}, which are more susceptible to movement due to their placement on the head. Studies leveraging \gls{hrv} typically identify drowsiness patterns by applying time and frequency domain analysis to the series of RR intervals (i.e. peak-to-peak time difference), by extracting different well-known metrics such as SDNN, RMSSD, and pNN50, for the time-domain, and LFHF ratio from the \gls{lf} and \gls{hf} bands~\cite{Awais2017, task1996heart}. While these studies focus specifically on \gls{ecg}-based drowsiness detection, many do not explicitly report window-shifting approaches to analyze the temporal trends of 
\gls{hrv} features. Among the studies that discuss \gls{ecg} window segmentation, Maftukhaturrizqoh et al. \cite{olivia_maftukhaturrizqoh__2019} used 30-second intervals, and Rachamalla et al. \cite{c__santhosh_kumar_2023} adopted 1-minute window segments. Similarly, Arefnezhad et al. \cite{sadegh_arefnezhad__2020} employed a 40-second sliding window with a 30-second overlap, Xiong et al. \cite{hui_xiong__2023} applied 1-minute windows with a 5-second step, and Xia et al. \cite{yan_wang__2019} segmented the signal into 30-second non-overlapping granules. However, most of these studies do not explicitly address the optimization strategies behind their chosen window-shifting parameters. This oversight limits the development of robust, data-driven models for drowsiness detection. It also highlights the need for a systematic approach to parameter tuning, such as the multi-configuration benchmarking methodology proposed in \cite{9161259} on which we base our strategy for capturing the optimal feature trends for drowsiness detection.

\subsection{Deep Reinforcement Learning and Braking Strategies}
\label{subsec:deep_q_networks_and_braking_strategies}

The advantage of \gls{drl} over supervised learning for adaptive braking strategies lies in its ability to explore and exploit behaviors through direct interaction with the environment. Supervised learning, on the other hand, is limited to pre-labeled datasets and struggles with unstructured or dynamic environments. In the context of \gls{drl}, an agent observes the current state of the environment, selects an action, and receives a reward signal that guides its learning process. The agent’s objective is to learn a policy that maximizes cumulative rewards. The \Gls{dqn} is a widely used \gls{drl} architecture that employs a neural network, referred to as the policy network, to approximate the Q-function used for evaluating actions. A core concept in \gls{dqn} is the Q-value function \( Q(s, a) \), which estimates the expected reward for taking action \( a \) in a state \( s \). It is updated iteratively using the Bellman equation~(\ref{eq:target_q_value}), where the target value \( y \) is defined as:
\begin{equation}
y = r + \gamma \max_{a'} Q(s', a')
\label{eq:target_q_value}
\end{equation}
where \( r \) is the immediate reward, \( \gamma \) is the discount factor, which determines how much the agent values future rewards, and \( s' \) and \( a' \) are the next state and action, respectively. The network parameters are then optimized by minimizing the mean squared error between the predicted Q-value and the target, using the loss function:
\begin{equation}
\mathcal{L} = \left( y - Q(s, a) \right)^2
\label{eq:loss_function}
\end{equation}

This mechanism, with one neural network, enables action evaluation but leads to instability due to rapidly changing Q-values. To address this issue, Mnih et al.~\cite{Mnih2015HumanlevelCT} proposed the use of a \gls{fdqn} approach, commonly referred to as the standard \gls{dqn}, which decouples the target from the rapidly fluctuating estimates and delays its update to stabilize learning. This is achieved by employing two neural networks: a policy network that estimates Q-values and a target network that is periodically updated with the weights of the policy network, thereby promoting training stability. A fundamental component in \gls{dqn} is the replay memory that stores past experiences as transitions. The memory enables random batch selection during training, which helps prevent bias from consecutive states. During training, action selection is guided by exploration mechanisms. One common mechanism in the \gls{dqn} agent's training follows an epsilon-greedy strategy, where the probability of selecting a random action depends on \( \epsilon \); This parameter decays over training episodes to allow a gradual transition from exploration to exploitation~\cite{9576818}. Even when a seemingly optimal strategy is found, maintaining a small level of exploration through the \(\varepsilon\)-greedy mechanism allows the agent to discover alternative strategies that may yield equal or better performance, thereby improving policy robustness. Thus, while the \(\varepsilon\)-greedy strategy is widely used, it may be inefficient in exploring optimal policies due to its uniformly random action selection~\cite{Ruffini_2016}. Nonetheless, this stochasticity serves as a necessary guidance mechanism in the early stages of training, enabling the agent to discover its first optimal strategy by preventing premature convergence to suboptimal behaviors. Preference-guided exploration can complement \( \epsilon \)-greedy by adjusting the probability distribution over actions during the early training phase, which increases chances of exploring initial behavior such as acceleration. This guidance improves learning efficiency while preserving exploration diversity and supporting better generalization. Several enhancements to \Gls{dqn} have been proposed to improve learning stability and overall performance. Among these, one notable method is the \gls{dodqn} algorithm, which addresses the overestimation bias commonly observed in \gls{dqn} by decoupling the action selection and evaluation processes~\cite{10.5555/3016100.3016191}. In standard \gls{dqn}, the target value is computed using the same network to both select and evaluate the next action, which can lead to overestimated Q-values. In contrast, \gls{dodqn} modifies the target value by decoupling action selection and evaluation:
\begin{equation}
y^{\text{DDQN}} = r + \gamma Q_{\text{target}}\left(s', \arg\max_{a'} Q_{\text{policy}}(s', a')\right)
\label{eq:ddqn_target}
\end{equation}
where \( Q_{\text{policy}} \) and \( Q_{\text{target}} \) denote the Q-value predictions for the next state \( s' \), produced by the policy and target networks, respectively. This separation mitigates overestimation bias, resulting in more accurate value updates and improved convergence. The loss function remains the same as in standard \gls{dqn}, typically defined as the mean squared error between the predicted Q-value \( Q(s, a) \) and the target value \( y^{\text{DDQN}} \). Another advancement is the \gls{dudqn} architecture~\cite{10.5555/3045390.3045601}, which separates the evaluation of state values from action advantages. This design is particularly effective in scenarios where small variations in actions significantly impact performance, such as small variations of braking pressure to evade collisions. In \gls{dudqn}, the \( Q(s, a) \) is decomposed into two separate estimators: the state value function \( V(s) \) and the advantage function \( A(s, a) \), combined as:
\begin{equation}
Q(s, a) = V(s) + \left( A(s, a) - \frac{1}{|\mathcal{A}|} \sum_{a'} A(s, a') \right)
\label{eq:dudqn_q_value}
\end{equation}
Both \( V(s) \) and \( A(s, a) \) are learned from a shared state representation, with the network separated into distinct outputs for value and advantage.

The applications of \Gls{dqn} for braking strategies have aimed to approximate optimal braking levels. For instance, Chae et al. \cite{8317839} utilized a \Gls{dqn} framework to optimize \gls{ttc} for pedestrian scenarios, reporting a 0\% collision rate for \gls{ttc} above 1.5s. Similarly, Liu et al.~\cite{9097944} proposed a \gls{dqn}-based framework for trajectory planning and braking optimization in \gls{v2i}-assisted autonomous driving, aiming to improve both driving safety and fuel efficiency through informed decision-making; They also utilize \gls{dodqn} variant within the same context, reporting notable improvements in both safety and fuel economy compared to the baseline \gls{dqn} model. For instance, a study by Liao et al. \cite{9190040} used \gls{dudqn} to learn highway environments and driving strategies while incorporating braking. Additionally, Hoel et al. \cite{8911507} combined planning and \gls{drl} for tactical decision-making. Similarly, Isele et al.~\cite{DBLP:journals/corr/IseleCSF17} employed a \gls{dqn} framework which was enhanced by a \gls{dudqn} architecture to develop safe navigation strategies at intersections lacking traffic signals, where braking is critical to avoid collisions. Makantasis et al.~\cite{doi:10.1049/iet-its.2019.0249} proposed a \gls{dodqn}-based driving policy for highway scenarios, achieving 0\% collision rate under rule-constrained settings, while in unconstrained configurations, the policy exhibited a 2–4\% collision rate, emphasizing a trade-off between safety and generalization in uncertain environments. This performance improvement is attributed to the use of \gls{dodqn}, which mitigates Q-value overestimation and improves decision-making stability in complex driving conditions. More recently, Yousaf et al.~\cite{YOUSAF2024112320} made a valuable contribution by introducing a stochastic rule-based speed adjustment mechanism within an \gls{eeg}-informed \gls{drl} framework, demonstrating the potential of integrating physiological signals into autonomous decision-making. However, the speed adjustments are predefined and not learned through interaction with vehicle dynamics, traffic conditions, and driver physiology. As a result, the decoupled structure, which separates cognition from control, limits the agent’s ability to adaptively learn context-aware  control behavior which highlights the need for an integrated learning framework.

Previous studies primarily focused on optimizing vehicle braking through \gls{drl} but did not account for the driver's physiological state. Given that drowsiness increases reaction times and impairs responsiveness, neglecting such physiological factors in braking strategies poses significant safety risks. To bridge these research gaps, we train, evaluate, and integrate an \gls{ecg}-based drowsiness detection \gls{rnn} model into a DQN framework that leverages dueling and double \gls{dqn} architectures for enhanced performance. These advancements enable dynamic braking strategies that adapt to driver state, improving both collision avoidance and overall driving safety.

\section{Proposed Drowsiness-Aware Braking System}
\label{sec:Proposed System}
In this section, we first present the architecture design of our proposed drowsiness-aware braking system. Then, we formalize the problem for the \gls{drl} approach. 

\subsection{System architecture}
\begin{figure}[t]
  \centering
  \includegraphics[width=\linewidth]{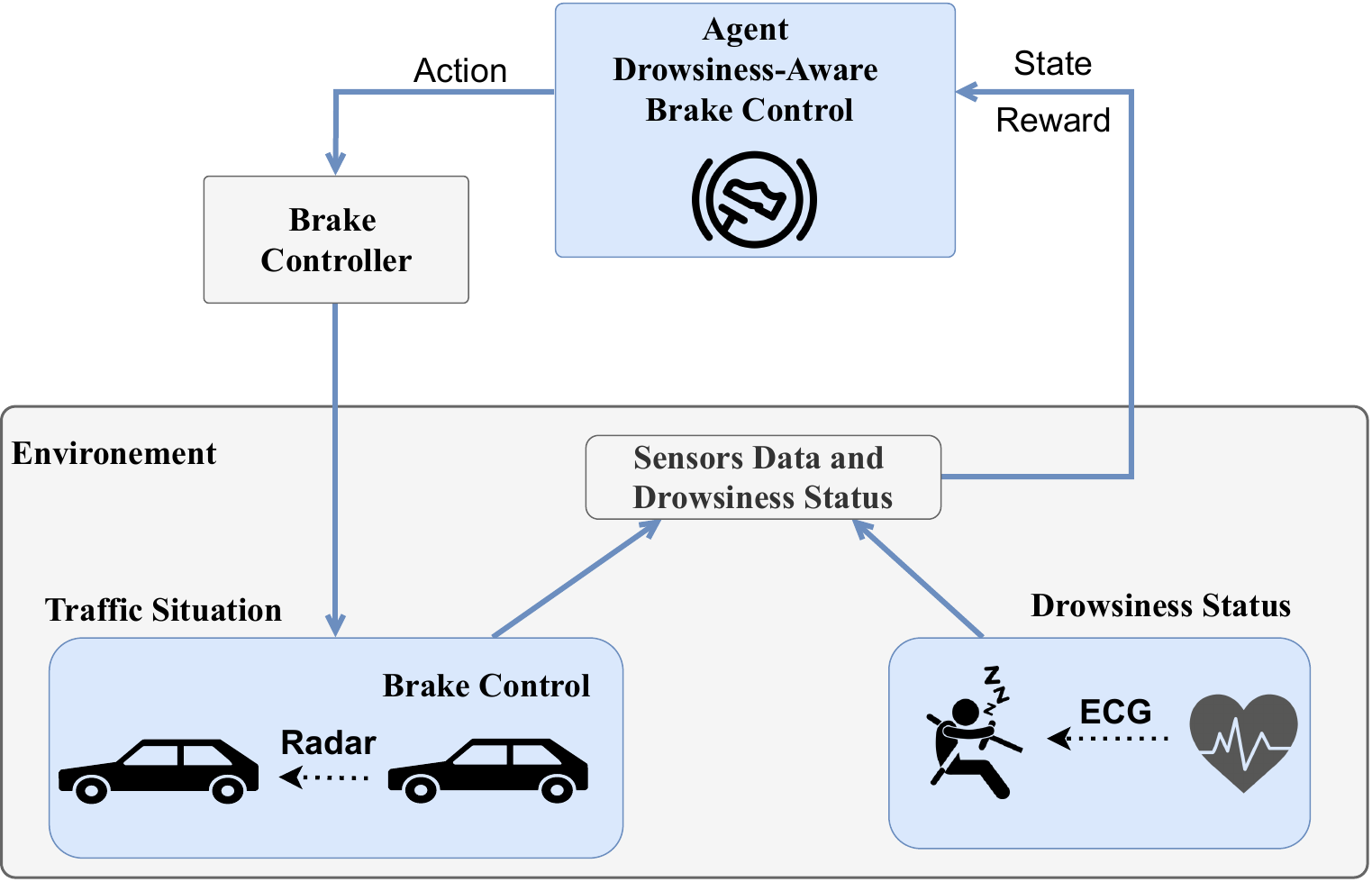}
    \caption{System architecture design for real-world drowsiness-aware brake control.}
    \label{fig:scenario_carla}
\end{figure}

The proposed system is designed as a modular, real-world braking architecture that adapts vehicle behavior based on physiological and contextual input. Figure~\ref{fig:scenario_carla} illustrates the overall system design, highlighting the interaction between sensory input, drowsiness detection, decision-making, and control execution. At its core, a \gls{dqn}-based agent receives information from the vehicle onboard sensors and a physiological monitoring module to assess the driving context and adjust braking decisions accordingly. The three main inputs to the agent include:
\begin{itemize}
    \item Drowsiness status inferred from ECG-derived features,
    \item Environmental context from radar and traffic conditions,
    \item Current braking state and vehicle dynamics.
\end{itemize}
Drowsiness is detected via \gls{hrv} features extracted from an \gls{ecg}, which reflect \gls{ans} activity. This choice is motivated by the availability of commercial wearable devices, such as chest bands, capable of providing \gls{ecg} and \gls{rri} signals for real-time analysis. When drowsiness is detected, the agent accounts for the delayed driver responsiveness by adapting its control strategy, favoring smoother and earlier braking to mitigate collision risk. The system incorporates a dedicated brake controller module that translates the agent’s decision into actuation commands.
The traffic situation is observed using the front radar mounted on the ego vehicle, which refers to the vehicle controlled by our braking agent. This radar collects velocity and depth information when a front vehicle is detected. As demonstrated in Chae et al.~\cite{8317839}, braking decisions primarily rely on velocity and depth. Therefore, only these two radar features are used for decision-making. To reduce noise and the dimensionality of the radar detections, the \gls{dbscan} algorithm \cite{Su202443139} is employed. \gls{dbscan} improves radar reliability by filtering out sensor noise and reflections. It organizes detections into dense clusters, enabling the \gls{dddqn} agent to operate on cleaner, more structured inputs for braking decisions.

\subsection{Formalization of the Problem}
\label{subsec:Problem Formalisation}
The agent’s objective is to maintain a safe stopping distance from the front vehicle, defined as \( d_{\text{stop}} \), to avoid collisions by adjusting the vehicle control inputs (i.e., braking pressure) while accounting for slower responses caused by a drowsy state. We model the drowsiness-aware adaptive braking agent as a \gls{mdp}. The \gls{mdp} is represented with the 5-tuple\( \langle S, A, P, R, \gamma \rangle \), where:

\begin{itemize}
    \item \( S \) is the set of states.
    \item \( A \) is the set of actions.
    \item \( P\) is the state transition probability function.
    \item \( R(s, a) \) is the reward function.
    \item \( \gamma \in [0,1] \) is the discount factor.
\end{itemize}

The state \( s_t \) at time \( t \) encapsulates all necessary information for decision-making, including vehicle state, environmental state, and driver state. It is represented as:

\begin{equation}
s_t = \left[ v_t, a_t, d_{\text{rel},t}, v_{\text{rel},t}, \theta_t \right]
\label{eq:state_space}
\end{equation}

In this state vector: 
\begin{itemize}
    \item \( v_t \) denotes the current speed in \( \mathrm{m/s} \).
    \item \( a_t \) is the current selected action.
    \item \( d_{\text{rel},t} \) is the radar's relative depth (i.e., distance to the front vehicle), computed as the centroid of clusters obtained via \gls{dbscan}.
    \item \( v_{\text{rel},t} \) is the radar's relative velocity, calculated similarly to \( d_{\text{rel},t} \). 
    
    \item \( \theta_t \) indicates the drowsy state, predicted by our proposed drowsiness classification model (\( \theta_t = 1 \) signifies active drowsiness).
\end{itemize}

Formally, let $\mu(C_i)$ denote the centroid of cluster $C_i$. The representative cluster point $p_c$ is computed as:
\begin{equation}
p_c = \frac{1}{k} \sum_{i=1}^{k} \mu(C_i)
\label{eq:average_membership}
\end{equation}

where $\mu(C_i)$ is the centroid of cluster $C_i$, calculated as:
\begin{equation}
\mu(C_i) = \frac{1}{|C_i|} \sum_{p \in C_i} p
\label{eq:cluster_mean}
\end{equation}
This approach ensures that the final detection point accurately reflects the central position of the most relevant clusters. If the radar returns only one detection point (\( n = 1 \)), that point is directly used without clustering. This centroid-based representation provides a compact and reliable representation of the lead vehicle, thereby reducing the state space complexity.

Following the discretization strategy proposed by Chae et al.~\cite{8317839}, the action space \( A \) is constructed using fixed pressure levels. This structure aligns with the \gls{dqn} framework and supports stable policy learning through interpretable control signals. The action \( a_t \) in \( s_t \) represents the control input at time \( t \), which may correspond to braking, acceleration, or a neutral command. It is discretized into a finite set of possible actions, formally defined as:

\begin{equation}
A = \{ a_0, a_1, a_2, a_3, a_4, a_5 \}
\label{eq:action_space}
\end{equation}
where each action corresponds to a specific control input as follows:

\begin{itemize}
    \item \( a_0 \): Braking with 100\% pressure.
    \item \( a_1 \): Braking with 70\% pressure.
    \item \( a_2 \): Braking with 40\% pressure.
    \item \( a_3 \): Braking with 20\% pressure.
    \item \( a_4 \): Neutral (no braking or acceleration).
    \item \( a_5 \): Full Acceleration with 100\% pressure.
\end{itemize}
Note that \( a_5 \) is the only action that enables forward motion for exploration purposes. The remaining actions are dedicated to braking control. 

The transition function in the \gls{mdp} determines the probability \( P(s_{t+1} | s_t, a_t) \) of transitioning to the next state \( s_{t+1} \) given the current state \( s_t \) and action \( a_t \). In model-free \gls{drl} methods like \gls{dqn}, this probability is unknown and not explicitly modeled; instead, the agent learns how to act effectively through trial-and-error interactions, without learning the detailed behavior of the environment. The proposed framework enables the agent to engage with a control system driven by real-world vehicle behavior, enabling it to learn braking policies suitable for real-world deployment. Drowsiness effects (\(\theta_t\)) are incorporated into the transition process to influence how states change and guide policy learning under impaired conditions. This forces the agent to adapt its decisions under different drowsiness conditions. When \( \theta_t = 1 \), a temporal delay \( \Delta t \) is applied to represent increased driver reaction time \( d_{\text{reaction}} \), as reported in the empirical study \cite{Goel2009}. For this implementation, the minimum delay of \( \Delta t = 0.5 \) seconds is adopted.

The reward function in the \gls{mdp} is designed to encourage safe and efficient driving while penalizing unsafe actions, and is formally defined as:

\begin{equation}
\resizebox{\linewidth}{!}{$
R(s_t, a_t) =
\begin{cases} 
    -\alpha \cdot \mathbf{1}(\text{collision}), & \text{(collision penalty)} \\ 
    -\beta \cdot |\Delta p_t|, & \text{(abrupt pressure change penalty)} \\ 
    -\kappa \cdot \mathbf{1}(d_t < d_{\min}), & \text{(unsafe distance penalty)} \\ 
    +\delta \cdot \mathbf{1}(d_{\min} \leq d_t \leq d_{\max}), & \text{(reward for safe distance)} \\ 
    +\epsilon \cdot \mathbf{1}(\text{smooth braking}), & \text{(reward for smooth braking)}
\end{cases}
$}
\label{eq:reward_function}
\end{equation}

The agent learns to adjust braking behavior to maximize cumulative reward, guided by penalties and rewards. The rewards parameters \( \alpha, \beta, \kappa, \delta, \epsilon \) are constant weights that determine penalties and rewards associated with each driving behavior (c.f. Section~\ref{subsec:reward-function}). The indicator function \( \mathbf{1}(\cdot) \) evaluates to 1 when the specified condition is met and 0 otherwise. The term \( \Delta p_t \) represents the change in braking pressure at time \( t \), and \( d_t \) is the current following distance. The safe distance encouraged by the rewards is set to \( d_{\min} = \max(5, 2 v_t) \), which is the minimum safe following distance based on the widely accepted two-second rule~\cite{MICHAEL200055}, with \( d_{\max} = d_{\min} + 10 \) introducing a buffer margin.


The discount factor \( \gamma \in [0,1] \) in the \gls{mdp} determines the importance of future rewards in decision-making. In this work, \( \gamma = 0.9 \), following established literature~\cite{8317839}, to ensure the agent considers long-term safety when selecting braking actions. Thus, the objective of the agent designed as \gls{mdp} is to find an optimal policy \( \pi^* \) that maximizes the expected cumulative reward:

\begin{equation}
\pi^* = \arg\max_{\pi} \mathbb{E} \left[ \sum_{t=0}^\infty \gamma^t R(s_t, \pi(s_t)) \right]
\label{eq:policy_optimization}
\end{equation}

The stopping distance equation \( d_{\text{stop}} \) acts as an implicit objective, not enforced through hard constraints but learned through the agent’s reward-guided interactions with its environment. This \gls{mdp}-based formulation provides a solid foundation for the \gls{dddqn} agent to learn optimal braking strategies under varying cognitive states.


\section{Drowsiness detection via ECG}
\label{sec:Feature extraction}

As mentioned in Section~\ref{sec:Proposed System}, part of the system detects the driver's drowsy state to enhance the \gls{drl}-based braking strategies. This section describes our drowsiness detection via \gls{ecg} signal, specifically using \gls{hrv} features. We classify the data into normal sinus rhythm (non-drowsy) and drowsy states, which we label as \gls{nsrw} and \gls{dew}. Since datasets with drowsy events are usually unbalanced, we select one window of \gls{nsrw} for each \gls{dew} in the data. To extract the features, an ``inner'' window-shifting optimization approach was applied to the \gls{ecg} signals, adapting the technique presented in \cite{9161259}. Our method divides each analyzed window into different configurations and selects the optimal combination of detection model and configuration, based on cross-validation to ensure robustness across different data partitions.

\subsection{Extraction of DEWs and NSRWs}

\begin{figure}[t]
    \centering
    \includegraphics[width=\linewidth]{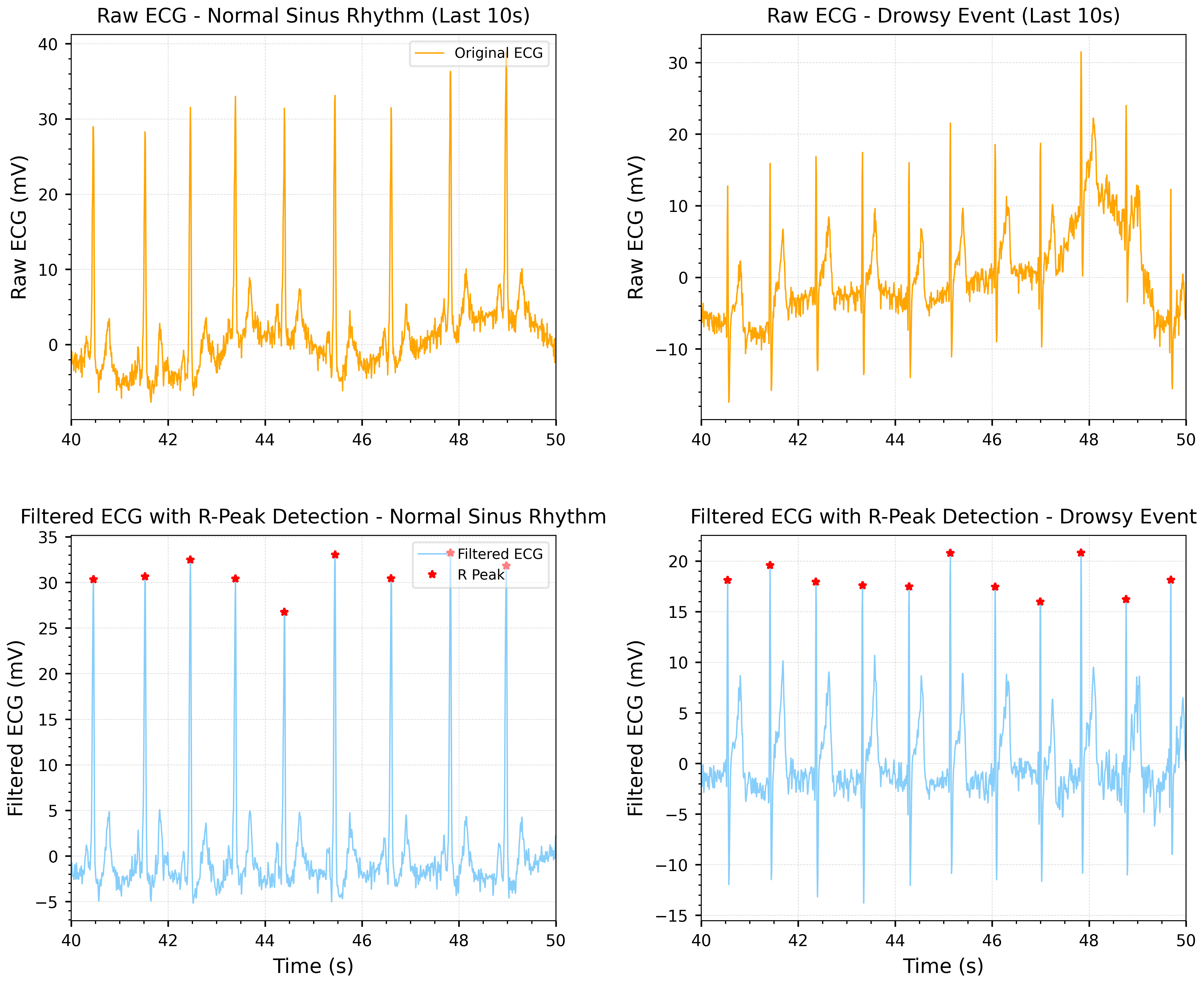}
    \caption{Raw and filtered ECG signals with R-peak detection for both normal sinus rhythm and drowsy event. The last 10 seconds of an ECG window from configuration C6400 N6 M72 are plotted. The first row presents the raw ECG signals, while the second row displays the filtered ECG signals with detected R-peaks.}
    \label{fig:ecg_r_peak_analysis_10s}
\end{figure}

\begin{figure}[t]
    \centering
    \includegraphics[width=0.9\linewidth]{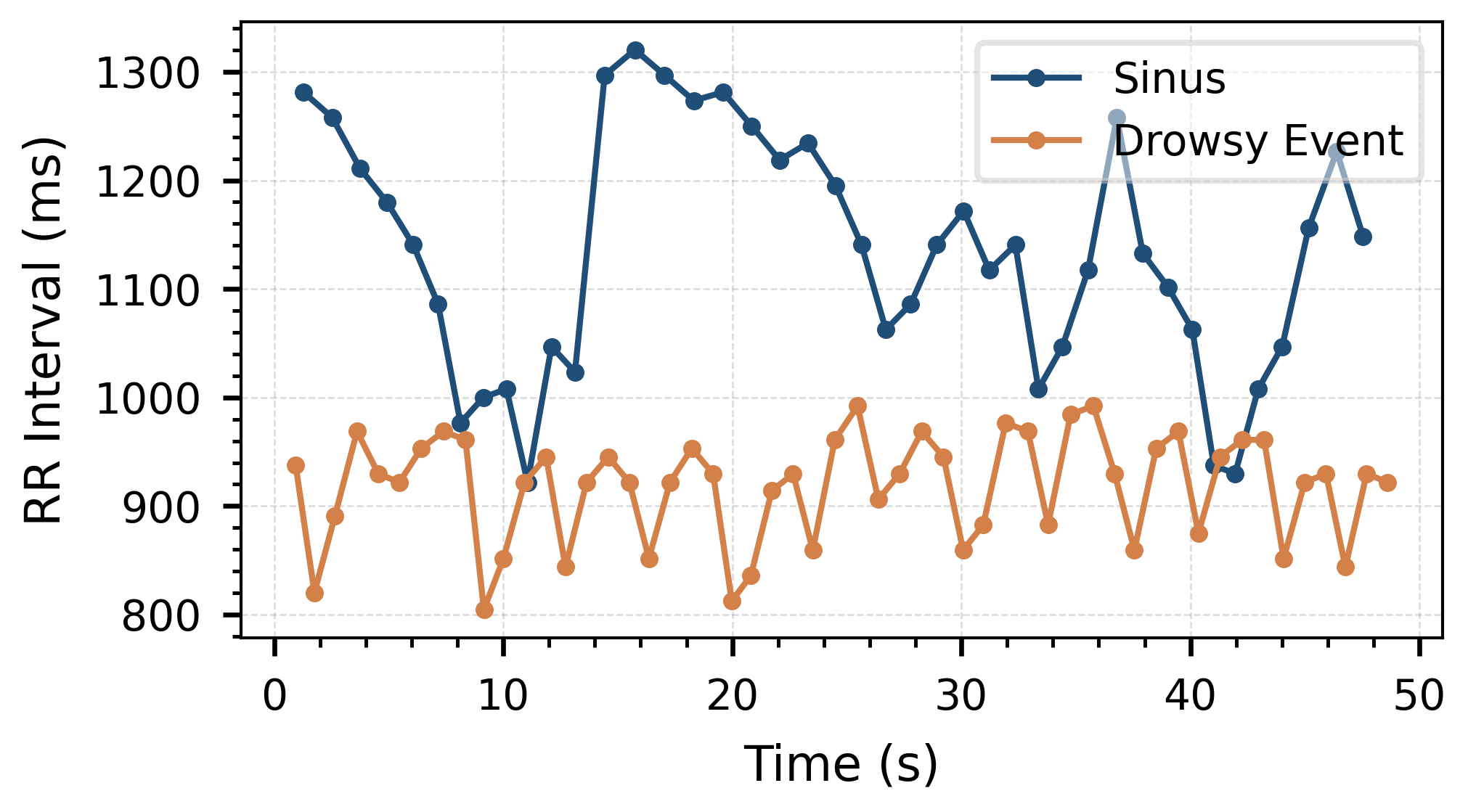}
    \caption{RR Interval trend over the last capsule of a random DEW and NSRW from C6400 N6 M72 configuration. The figure compares sinus rhythm (blue) and drowsy event intervals (orange), illustrating changes over time.}
    \label{fig:rr_trend}
\end{figure}

The drowsiness detection module begins with preprocessing the \gls{ecg} signal. First, we extract the \gls{dew}s that are far enough from each other, specifically at least two minutes between two drowsiness events~\cite{s24134316}. We also chose two minutes based on visual inspection and to ensure sufficient duration for reliable \gls{hrv} analysis, consistent with standard HRV practices (e.g., $>$60s for frequency-domain features~\cite{task1996heart}). Then, based on the chosen minimum duration and the method from~\cite{9161259}, we extract \gls{nsrw}s from the segments between two events, ensuring that \(\text{DEWs} \cap \text{NSRWs} = \emptyset\). \gls{nsrw}s are extracted using the same procedure as \gls{dew}s, with the condition that no drowsiness event occurs within the selected window. The second step is the \gls{ecg} analysis to extract relevant \gls{hrv} features from the resulting \gls{dew}s and \gls{nsrw}s. Figure~\ref{fig:ecg_r_peak_analysis_10s}~and~\ref{fig:rr_trend} shows an example of \gls{dew}s and \gls{nsrw}s. Specifically, we show the \gls{ecg} signal and the corresponding RR series, i.e., series of peak-to-peak intervals where the peak is commonly defined as "R peak". As expected, \gls{nsrw}s show a rhythm variability consistent with normal conditions at rest. In contrast, \gls{dew}s show patterns in the rhythm variability aligned with drowsiness-related physiological changes~\cite{Fujiwara20242956, s24134316}.

\subsection{Capsule-Shifting for HRV features extraction}
\begin{figure}[t]

    \centering
    \includegraphics[width=1\linewidth]{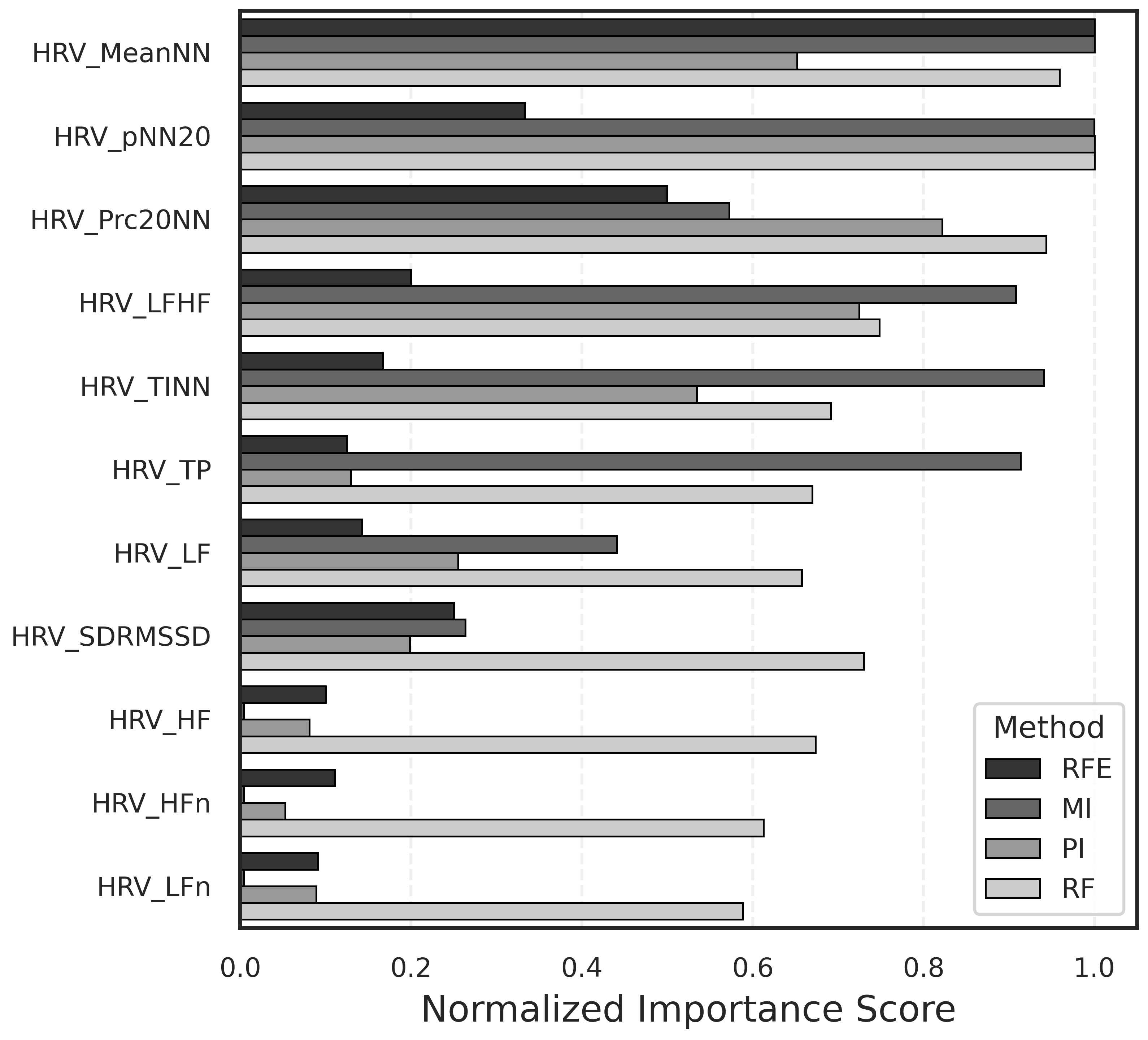} 
    \caption{Feature Importance Comparison Across Methods: RFE, MI, PI, and RF.}
    \label{fig:feature_importance}
\end{figure}

After extracting the windows, we divide them into shifting segments that we call "capsules". Capsule shifting is a \gls{tde} technique that reconstructs temporal dynamics by capturing sequential dependencies across overlapping segments. This technique is crucial for time-series models as it preserves the sequential nature of the data required to capture temporal dependencies while increasing the number of available samples for \gls{hrv}. The capsule configuration defines how \gls{dew}s and \gls{nsrw}s are segmented into sequences for the later \gls{hrv} analysis. In our approach, which we call CNM, we define each configuration using three parameters: the capsule size \(C\), expressed in number of samples; the capsule count \(N\), which indicates the number of capsules per \gls{dew} or \gls{nsrw}; and the capsule overlap \(M\), expressed as a percentage of shared data between consecutive capsules. Additionally, we consider a window overlap (\(M'\)) used only during inference to apply a sliding mechanism for prediction across the full \gls{ecg} signal. Then, within each capsule, we extract and analyze a set of \gls{hrv} features to train the models. Figure~\ref{fig:feature_importance} summarizes the selected \gls{hrv} features, which include time-domain and frequency-domain metrics, following existing literature \cite{Fujiwara20242956, Saleem2023}. The extracted features are evaluated using four different methods: \gls{rfe}, \gls{mi}, \gls{rf}, and \gls{pi}. First, \gls{rfe} recursively eliminates the least important features to identify the most relevant subset for classification. Second, \gls{mi} quantifies the dependency between each feature and the target label by measuring shared information. Third, \gls{rf} assesses feature importance based on its contribution to classification accuracy in a tree-based model. Finally, \gls{pi} evaluates feature significance by randomly shuffling feature values and analyzing the impact on model performance. Figure~\ref{fig:feature_importance} illustrates the selected feature importance ranking across these four methods, which highlights the most discriminative features in drowsiness detection for our case study.

A configuration is considered a candidate if the capsules are optimally sized and positioned to fit within the window \(L\), i.e., \gls{dew} and \gls{nsrw}. This method maintains data integrity and prevents distortions that could affect feature extraction and model performance. In \textit{CNM}, the capsule-shifting process begins with the initialization of the search space, where \(N\_range\) is defined in the range of [2, 200] and \(C\_range\) in the range of [40, 120]. The lower limit of 40 seconds is chosen as the minimum window length required for \gls{hrv} frequency-domain feature extraction using \gls{nk2}~\cite{Makowski2021neurokit}. The upper limit of 120 seconds corresponds to the maximum duration of \gls{dew}s and \gls{nsrw}s considered for drowsiness detection. For each pair of capsule size \(C\) in samples and number of capsules \(N\), the overlap fraction \(M\) is computed using the expression:

\begin{equation}
M = \frac{C \times N - L}{(N - 1) \times C}
\label{eq:overlap_computation}
\end{equation}

To ensure that the capsules are evenly spaced and fully cover the window \(L\) without requiring padding or truncation, we only accept values of \(M\) in the open interval \((0, 1)\) that can be expressed with finite decimal precision (e.g., two decimal places). Once valid configurations are identified, we address the challenge of the high computational cost of calculating \gls{hrv} features due to the large number of configurations. We use parallel computation with shared cloud memory, which has significantly reduced the total processing time.



\subsection{Drowsiness Model Training: Recurrent Neural Network}
\label{subsec:Model Training}
Time-series classification tasks need models capable of capturing temporal dependencies. In this study, we evaluated several deep learning architectures including \gls{lstm}, \gls{gru}, \gls{rnn}, \gls{cnn}-\gls{lstm}, and \gls{ae}-\gls{lstm}. Our objective was to identify the most effective model and optimal capsule-shifting configuration for drowsiness detection from ECG signals. 

Feature values were standardized to zero mean and unit variance using Scikit-learn’s StandardScaler. Early stopping was applied throughout training to prevent overfitting and to ensure consistent evaluation conditions across models. This technique monitors validation performance and stops training once the metric stops improving for a predefined number of epochs, thus avoiding unnecessary computation and performance degradation. A broad hyperparameter search was conducted to optimize each architecture individually. Each model-configuration pair was evaluated using \(k\)-fold cross-validation followed by testing on a separate hold-out set, with final results reported as the average performance across all \(k\) trained models (c.f.~Table~\ref{tab:best-configurations}). This rotation helped assess generalization and avoid overfitting to specific data segments. Based on the cross-validation analysis, the two configurations C10240\_N2\_M50 and C6400\_N6\_M72 were selected for further comparison due to their consistently high performance with all models. In a refined evaluation step, we focused on \gls{rnn} and \gls{lstm} models, which achieved better performance and lower architectural complexity than \gls{cnn}-\gls{lstm}, \gls{ae}-\gls{lstm}, and \gls{gru}. The combination of \gls{rnn} and C6400\_N6\_M72, which achieved an average CV test accuracy of 92.4\% and average F1-score of 92.0\%, was selected for final refinement due to its stable performance and longer sequence window compared to the pair \gls{rnn} and C10240\_N2\_M50 (c.f Table~\ref{tab:best-configurations}). The choice of capsule number \(N = 6\) provided an effective trade-off between capturing sufficient temporal context and maintaining training efficiency. The finalized architecture consists of three recurrent layers with 40 \gls{rnn} units each, a dropout of 0.25, and an L2 regularization of 0.28 applied to the final dense layer. Training was performed for up to 100 epochs using the Adam optimizer with a learning rate of 0.002.

\begin{table*}[t]
\centering
\caption{Performance Comparison of Optimized Deep Learning Models for Drowsiness Detection Using Capsule-Shifting Configurations}
\label{tab:best-configurations}
\setlength{\tabcolsep}{12pt}
\begin{tabular}{llccccccc}
\toprule
Model & Configuration & Units & Dropout & Layers & Learning Rate & Batch Size & Accuracy & F1 Score \\
\midrule
\multirow{2}{*}{LSTM} 
    & C10240\_N2\_M50 & 100 & 0.20 & 1 & 0.001 & 80 & 0.9080 & 0.9105 \\
    & C6400\_N6\_M72  & 40  & 0.25 & 1 & 0.001 & 64 & 0.8734 & 0.8734 \\
\midrule
\multirow{2}{*}{GRU}  
    & C10240\_N2\_M50 & 50  & 0.25 & 2 & 0.002 & 64 & 0.9360 & 0.9380 \\
    & C6400\_N6\_M72  & 100 & 0.20 & 2 & 0.002 & 64 & 0.8924 & 0.8931 \\
\midrule
\multirow{2}{*}{RNN}  
    & C10240\_N2\_M50 & 150 & 0.15 & 2 & 0.002 & 80 & 0.9560 & 0.9569 \\
    & \textbf{C6400\_N6\_M72}  & \textbf{40}  & \textbf{0.25} & \textbf{3} & \textbf{0.002} & \textbf{48} & \textbf{0.9241} & \textbf{0.9200} \\
\bottomrule
\end{tabular}
\end{table*}

\section{Deep Reinforcement Learning Framework Based on DQN}
\label{sec:DQN based Deep RL Framework}

As described in Section \ref{sec:Proposed System}, our proposed system integrates multiple advanced modules to train a \gls{dddqn} agent while considering drowsiness. This approach applies a \gls{dodqn} and a \gls{dudqn} to train the agent. This enhances learning stability and yields more reliable Q-value estimates (cf. Section~\ref{sec:Background and Related Work}).

\begin{figure*}[t]
    \centering
    \includegraphics[width=\linewidth]{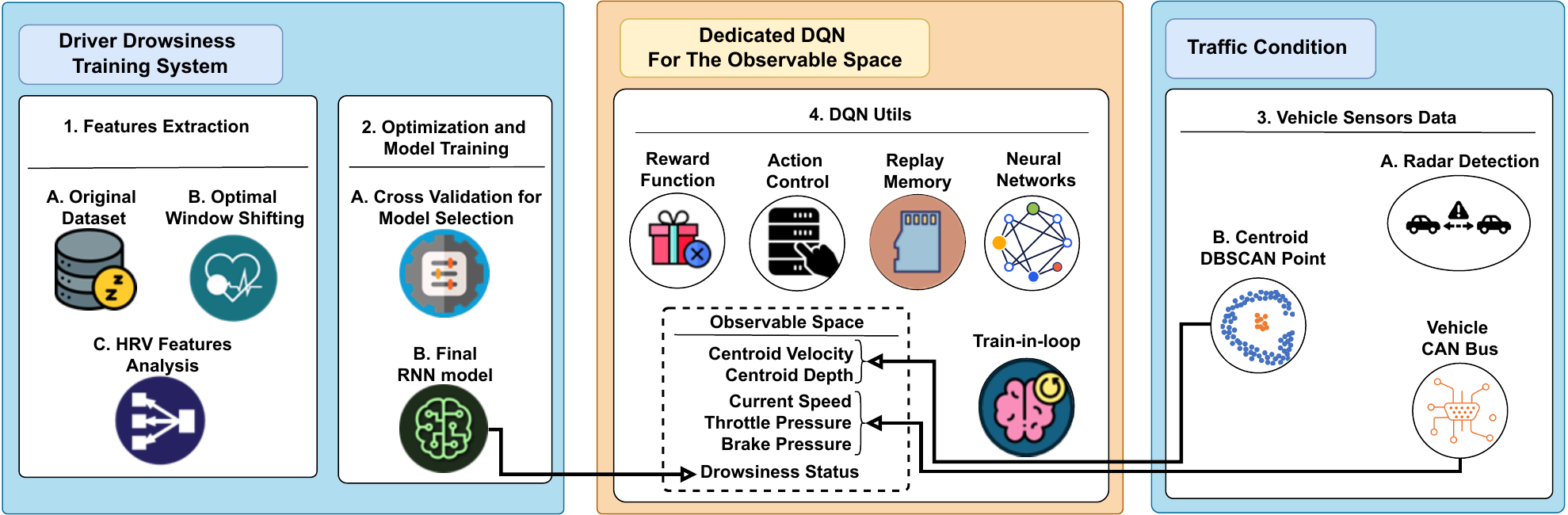}
    \caption{Overview of the proposed \gls{drl} framework, showing the interaction between the driver drowsiness detection system (left), the \gls{dqn} agent (center), and the vehicle sensor modules (right). The framework integrates HRV-based drowsiness predictions, radar-based front vehicle detections (filtered with \gls{dbscan}), and throttle, brake, and speed signals retrieved from the CAN bus to construct the observable state space. The agent uses this input to learn context-aware braking strategies under both alert and impaired cognitive conditions.}
    \label{fig:architecture7}
\end{figure*}

Figure~\ref{fig:architecture7} illustrates the complete \gls{drl} framework, comprising three main modules: the driver drowsiness detection system (left), the dedicated \gls{dqn} agent (center), and the traffic sensing module (right). The first block (Section~\ref{sec:Feature extraction}) extracts \gls{hrv} features from \gls{ecg} using optimized window shifting, with the best \gls{rnn} model selected via cross-validation to provide real-time drowsiness predictions. The third block (Section~\ref{sec:Proposed System}) gathers radar-based detections, filtered using \gls{dbscan}, to capture traffic context, along with the \gls{can} bus telemetry ( a standard vehicle communication protocol used to transmit real-time data such as speed, throttle, and brake pressure) that provides real-time vehicle dynamics. The center block receives this combined input, including centroid velocity, depth, throttle, brake, and drowsiness status, to construct the observable space. The agent then learns context-aware braking behavior through action control, reward shaping, and \gls{dqn} logic, which enables adaptive responses to impaired cognitive states.

\subsection{Step Function and Reward Structure}
\label{subsec:reward-function}

The step function is a key component of the proposed \gls{dqn} framework, as it handles state transitions, reward computation, and scenario termination. We design this function to integrate these processes to enable dynamic interactions between the agent and the environment. Upon selecting an action \( a_t \) from the discrete action space \( A \) defined in Eq. \ref{eq:action_space} in Section~\ref{subsec:Problem Formalisation}, the step function applies the selected action to the vehicle, transitioning the system from the current state \( s_t \) to the next state \( s_{t+1} \). If drowsiness is detected (\( \theta_t = 1 \)), the step function delays the selected action. The immediate reward \( r_t \) is then computed using the reward function defined in Eq. \eqref{eq:reward_function}. When the tuple \(\langle s_t, a_t, r_t, s_{t+1} \rangle\) is composed, a step is considered complete, and the tuple is appended to the replay memory which follows the common practice in \gls{dqn}  of using a maximum size of \(10^6\) transitions \cite{9576818}. A \gls{fifo} mechanism is applied when replay memory reaches \(10^6\) experiences, removing the oldest tuple to make space for a new transition. This ensures the agent learns from a continuously updated set of experiences and benefits from more recent transitions as its policy evolves. This interaction loop ensures that each action, including its potential delay under drowsiness, contributes to a meaningful experience tuple used during training. At each timestep \(t\), the agent executes an action \(a_t\) that interacts with the environment and modifies the following parameters of the so-called ``ego vehicle'' (i.e., the vehicle where the drowsiness-aware adaptive braking strategy is applied):
\begin{itemize}
    \item The ego vehicle's throttle, \( \tau_{\text{ego}} \), where \( \tau_{\text{ego}} \in [0, 1] \).
    \item The ego vehicle's braking pressure, \( \beta_{\text{ego}} \), where \( \beta_{\text{ego}} \in [0, 1] \).
\end{itemize}

\begin{figure*}[t]
    \centering
    \includegraphics[width=\linewidth]{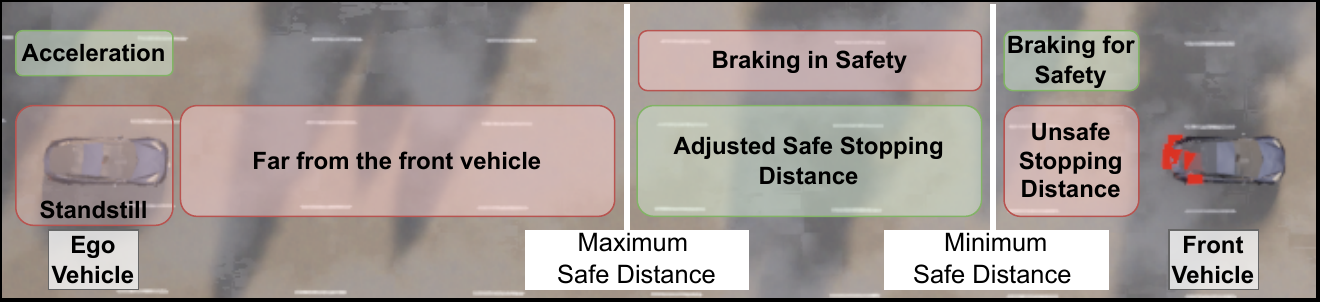}
    \caption{Reward structure visualization: green labels indicate positive rewards; red labels indicate penalties based on distance and braking behavior.}
    \label{fig:reward_structure}
\end{figure*}

The reward function \( r_t \), defined in Eq.~\ref{eq:reward_function} in Section~\ref{subsec:Problem Formalisation}, promotes safe driving by assigning positive feedback to desirable actions and penalties to unsafe behaviors. As shown in Fig.~\ref{fig:reward_structure}, the reward design is structured around three behavioral phases that guide the agent during \gls{dddqn} training. In the first phase, the agent learns to initiate movement from a standstill. This initial behavior is shaped by a reward signal that encourages forward acceleration while discouraging unnecessary braking or remaining stationary. The second phase focuses on maintaining a safe following distance behind the lead vehicle. This distance is dynamically calculated using a two-second rule formulation~\cite{MICHAEL200055}, and the agent is rewarded for staying within this adaptive range. Additional feedback encourages appropriate braking when the front vehicle is critically close and penalizes excessive closing speeds to promote smoother adaptation. The final phase handles critical safety violations. Collisions are treated as the most severe outcome and are assigned a high penalty to dominate the episode's cumulative reward. A small positive reward is granted for completing the scenario safely.

\begin{figure}[t]
  \centering
  \includegraphics[width=\linewidth]{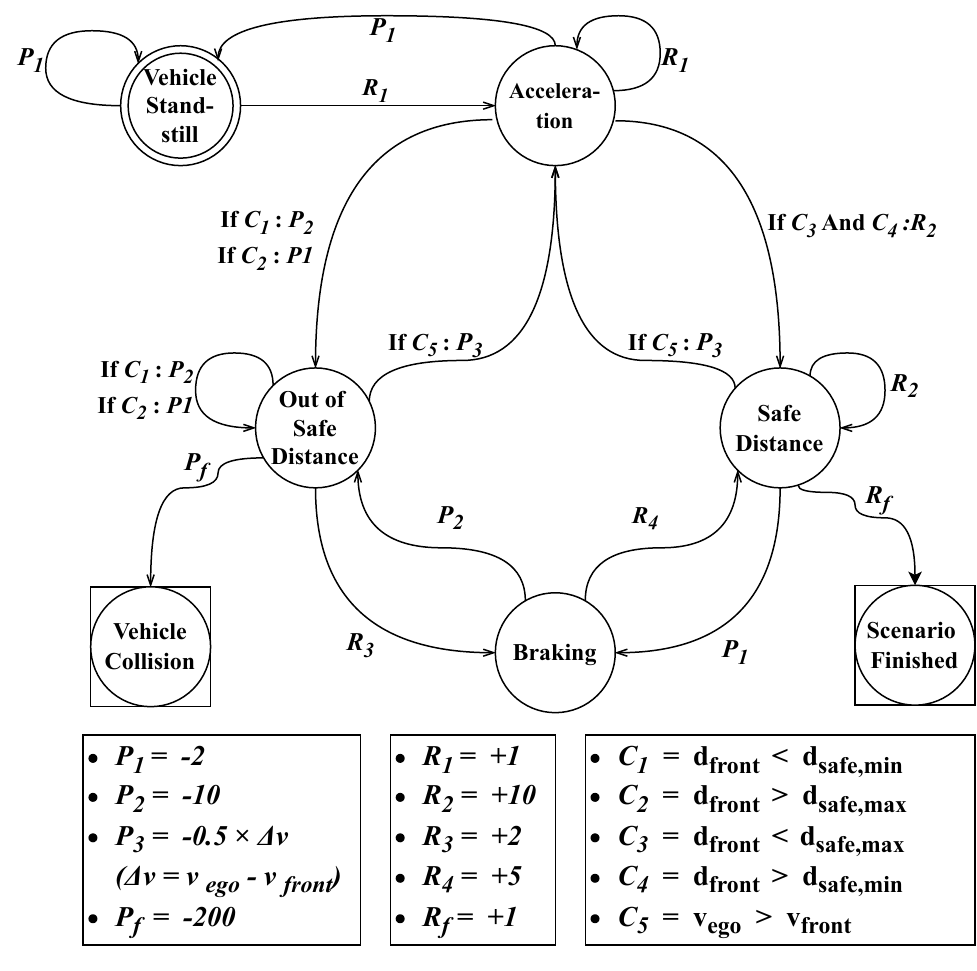}
  \caption{Finite state automaton representing the reward structure across the considered states during braking. P denotes penalty, R denotes reward, and C denotes condition related to speed and safe distance.}
  \label{fig:reward_automaton}
\end{figure}

To provide a structured overview of the agent’s reward feedback across different driving contexts, Fig.~\ref{fig:reward_automaton} presents a finite-state automaton representing the reward logic. Each node corresponds to a behavioral state (e.g., standstill, safe following, or collision), and the edges are labeled with the respective rewards or penalties. While the automaton does not capture the full complexity of the environment or state space encountered during training, it reflects the intended reward structure that shapes learning. The agent must ultimately adapt its strategy in response to dynamic environmental conditions, using this feedback to develop safe and effective driving behaviors.

\subsection{Double Dual DQN Training and Drowsiness Integration}
\label{subsec:dddqn_drowsiness}

An \gls{ecg}-based drowsiness detection system is integrated into the braking scenario to simulate the effects of driver drowsiness on vehicle control. This system processes \gls{ecg} data into windows and capsules, calculates \gls{hrv} features, and predicts the driver's drowsy state, as shown in the left block of Figure~\ref{fig:architecture7}. The detection system runs in a separate thread to enable continuous monitoring without disrupting other processes. A study shows that drowsy reaction times range from 500 milliseconds to 20 minutes \cite{Goel2009}. A 500-millisecond delay \(\Delta t = 0.5\) is chosen corresponding to the lowest drowsy reaction times, to assess whether the \gls{dqn} agent can detect and adapt to minimal impairment. When the drowsiness model detects a drowsy state, the last executed action is delayed by 0.5 seconds. The driver's drowsy state is added to the observable state in the \gls{dqn}, which also includes the parameters related to the ego vehicle and radar information (cf. Section~\ref{subsec:Problem Formalisation}). This allows the agent to learn the relationship between delayed responses, drowsiness, and detected vehicles. 

To evaluate the agent’s ability to adapt under impaired conditions, a \gls{dddqn} training strategy is applied. Initially, the agent lacks prior knowledge. Relying solely on an epsilon-greedy strategy (cf. Section~\ref{sec:Background and Related Work}) in early episodes significantly delays the learning of acceleration, which is the primary cause of early collisions. This is addressed by using a guided exploration~\cite{Ruffini_2016} for the first 50~episodes, encouraging the agent to select action~5 (acceleration) with an 80\% probability. Full exploitation (100\% acceleration) is avoided to maintain exposure to braking and neutral actions. This enables the agent to experience penalty-inducing crash scenarios, therefore, accelerates learning by reinforcing the consequences of high-risk decisions, leading to faster reward convergence. After the guided phase, action selection reverts to a standard epsilon-greedy strategy, enabling a gradual transition toward exploitation.

Each episode ends either upon collision or after a fixed duration of 30 seconds. During each step, the agent logs its experience tuple into a replay memory buffer. Training begins once the buffer reaches the minimum \(10^6\) threshold of samples. A minibatch of past transitions is then sampled to compute target Q-values based on the Bellman equation (cf. Eq.~\ref{eq:target_q_value}), using predictions from both the policy and target networks. Double DQN logic is applied: action selection is based on the policy network, while value estimation relies on the target network to reduce overestimation bias. The policy network is updated by minimizing the difference between the predicted Q-value for the selected action and the target Q-value, typically using a mean squared error loss. To ensure stable learning, a periodic synchronization of weights is performed from the policy to the target network after a fixed number of updates.
Three braking phases are defined based on \( v_{\text{rel},t} \) and \( d_{\text{rel},t} \):
\begin{itemize}
    \item ``acceleration'' is identified when \( v_{\text{rel},t} < 0 \),
    \item ``braking'' when \( d_{\text{rel},t} < d_{\text{safe,min}} \) and \( v_{\text{rel},t} > 0 \), and
    \item ``following'' when \( v_{\text{rel},t} \approx 0 \) while \( d_{\text{rel},t} \in [d_{\text{safe,min}}, d_{\text{safe,max}}] \).
\end{itemize}
Defining these three phases allows evaluation of the agent’s behavioral adaptation across different driving situations and cognitive conditions. In summary, the proposed \gls{dddqn} framework integrates vehicle sensing, physiological monitoring, and cognitive delay modeling to train an agent capable of adapting braking behavior to dynamic road and awareness conditions.

\section{Experiments and Results}
\label{sec:Experiments and Results}

This section presents the experimental evaluation of the proposed drowsiness detection system and adaptive braking strategy using a publicly available dataset and the \gls{carla} simulator \cite{Dosovitskiy2017}. After describing the experimental setup and the drowsiness dataset, we present the results of the optimal configuration-architecture evaluation for drowsiness detection through cross-validation and hyperparameter tuning. Next, we analyze the performance of the \gls{dddqn} agent in real-time drowsiness-aware driving scenarios. We then compare the performance of the proposed \gls{dddqn} agent against standard \gls{dqn}, \gls{dddqn}, and \gls{dodqn} variants to assess the benefit of each enhancement. Finally, we assess the system’s capability to adapt braking strategies under drowsy conditions.

\subsection{Experimental Setup}
\label{subsec:experimental_setup}

To validate the theoretical formulation outlined in Section~\ref{subsec:Problem Formalisation}, we instantiated the proposed system — comprising the adaptive braking strategy and drowsiness-aware agent — within a simulation environment designed to approximate real-world constraints while preserving control over behavioral variability. The \gls{carla} simulator was selected as a formal ground to verify whether the modeled \gls{mdp} (cf.~Section~\ref{subsec:Problem Formalisation}) enables the agent to operate effectively under realistic temporal and spatial conditions.

\begin{figure}[t]
    \centering
    \includegraphics[width=\linewidth]{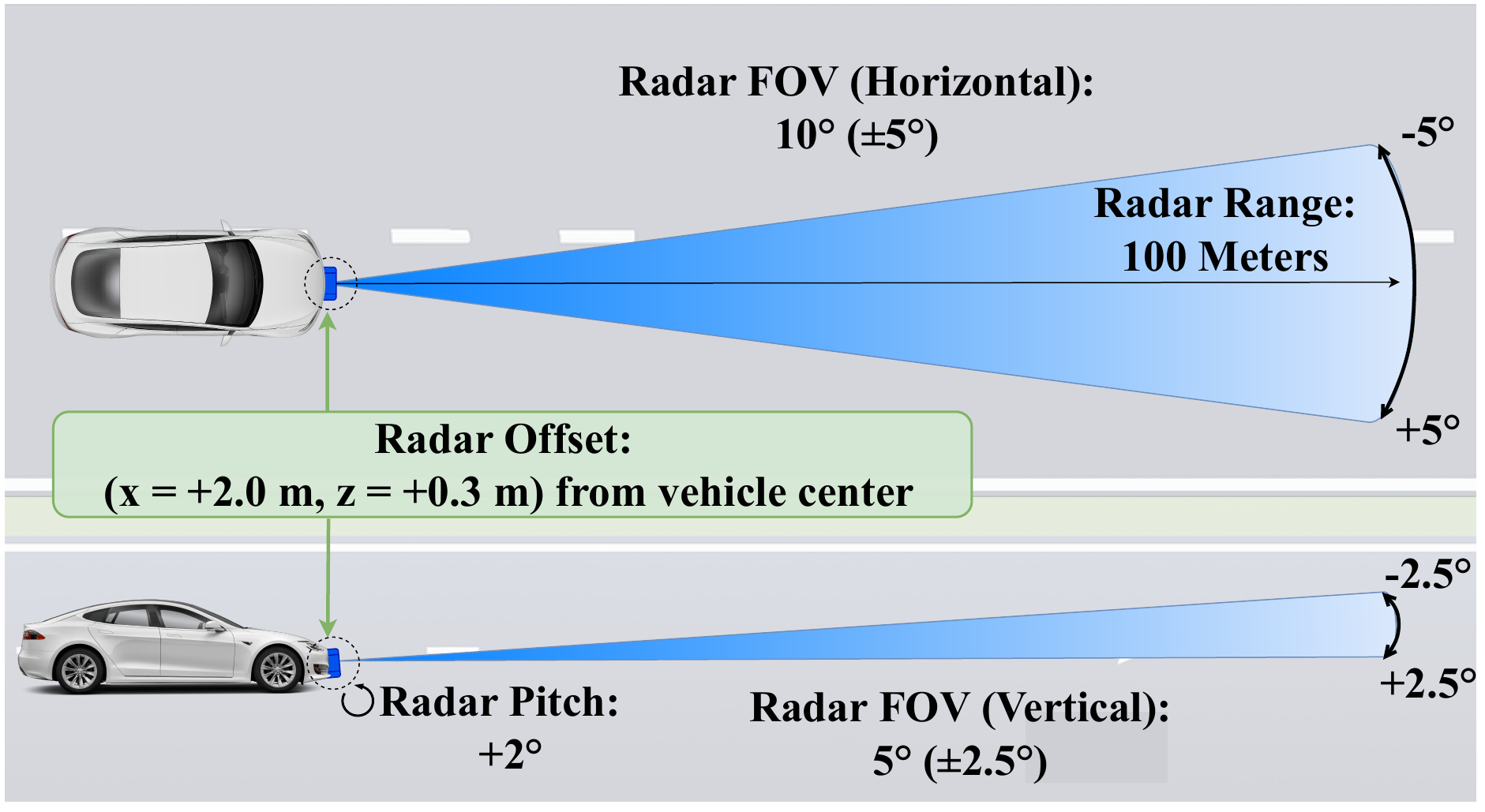}
    \caption{Front-mounted radar sensor showing forward field of view and detection zone.}
    \label{fig:radar_placement}
\end{figure}

Figure~\ref{fig:radar_placement} illustrates the radar sensor placement and field of view used in the simulation of the drowsiness-aware autonomous braking scenario. A front-facing radar sensor is mounted on the ego vehicle’s front bumper. The radar’s pitch angle is set to +2°, enabling slightly elevated forward perception in line with road geometry. This sensor continuously captures distance, velocity, and angular position of detected objects with up to 1000 detection points per second. To extract meaningful lead-vehicle estimates, radar detections are clustered using the \gls{dbscan} algorithm.

The agent's observable state vector in the simulation corresponds exactly to the formal state definition \( s_t = [v_t, a_t, d_{\text{rel},t}, v_{\text{rel},t}, \theta_t] \), combining ego speed, selected control action, radar-derived measurements, and driver drowsiness state (cf. Section~\ref{subsec:Problem Formalisation}). This state is continuously updated throughout each step of an episode and serves as input to the agent's action selection process. The action space \( A \) is discretized into six levels, each corresponding to a distinct actuator command:
\begin{itemize}
    \item \( a_0 \): Full braking (brake = 1.0),
    \item \( a_1 \): Strong braking (brake = 0.7),
    \item \( a_2 \): Moderate braking (brake = 0.4),
    \item \( a_3 \): Light braking (brake = 0.2),
    \item \( a_4 \): Coasting (throttle = 0.0, brake = 0.0),
    \item \( a_5 \): Full acceleration (throttle = 1.0, brake = 0.0).
\end{itemize}
These discrete control signals are mapped to throttle and brake pressure controls, applied to the simulated vehicle.

Each simulation episode runs for 30 seconds or terminates prematurely in the event of a collision. It is configured with random initial conditions such as vehicle spacing and throttle settings, which are randomized to ensure generalization of the learned policy. This setup provides a variable environment that mirrors real-world complexities while supporting rigorous evaluation of braking behavior under cognitive impairment. The setup instantiates the proposed \gls{mdp}, integrating perception, cognitive state, and delayed responses to train safety-aware braking policies.

\subsection{Drivers drowsiness database}
\label{sec:drowsiness_dataset}

We used the public \gls{dddb} \cite{Orosco_Lorena_2023}, which contains a comprehensive collection of physiological signals recorded from 10 healthy volunteers aged 20 to 50 years. The dataset was collected during simulated driving experiments designed to induce drowsiness under controlled conditions. The driving simulator experiments were conducted in a quiet, dark environment simulating a nighttime driving scenario on a straight route. Volunteers, partially sleep-deprived the night before, performed two 2-hour driving trials on separate days. This protocol ensured controlled drowsiness induction while collecting approximately 4 hours of physiological data per volunteer. Volunteers were instructed to press an event button upon feeling drowsy, and these annotations serve as reliable ground truth markers for drowsiness detection.

The dataset includes two 2-hour-long \gls{ecg} recordings with labels marking drowsiness events identified when volunteers pressed an event button during simulated driving. The \gls{ecg} signals were acquired using a bipolar configuration in the DIII lead position with a sampling frequency of 128 Hz and a 16-bit resolution that accurately captures heart activity sensitive to drowsiness. Unlike preprocessed datasets, the \gls{dddb} provides raw \gls{ecg} recordings, allowing for custom preprocessing and feature extraction. Moreover, it includes real-time drowsiness annotations, unlike datasets that label entire sessions as uniformly drowsy or non-drowsy. 

During data preprocessing and training, a marked imbalance between \gls{dew} and \gls{nsrw} introduced a significant risk of overfitting. Although \gls{hrv} features were extracted from all \textit{CNM} configurations, the classifier consistently favored the majority class. To mitigate this bias, we applied a class balancing strategy that preserved an equal number of non-drowsy samples for each drowsy event, ensuring a representative and unbiased training set without overrepresenting either class. To integrate all components seamlessly, the entire pipeline, including data processing, training, and simulation control, was implemented in Python and integrated with the \gls{carla} simulator through its Python 3.8.10\gls{api}.

\subsection{Drowsiness detection performance}
\begin{table*}[ht]
\centering
\setlength{\aboverulesep}{0.2em}
\setlength{\belowrulesep}{0.2em}
\caption{The average test of the first cross-validation results for comparison of capsule-Shifting configurations and deep learning models for drowsiness detection.}
\label{tab:cs_model_comparison}
\resizebox{\textwidth}{!}{%
\begin{tabular}{@{\hspace{6pt}}l@{\hspace{8pt}}cc@{\hspace{4pt}}cc@{\hspace{4pt}}cc@{\hspace{4pt}}cc@{\hspace{4pt}}cc@{\hspace{6pt}}}
\toprule
Configuration & \multicolumn{2}{c}{\textsc{Rnn}} & \multicolumn{2}{c}{\textsc{Cnn-Lstm}} & \multicolumn{2}{c}{\textsc{Ae-Lstm}} & \multicolumn{2}{c}{\textsc{Lstm}} & \multicolumn{2}{c}{\textsc{Gru}} \\
& \multicolumn{2}{c}{\hrulefill} & \multicolumn{2}{c}{\hrulefill} & \multicolumn{2}{c}{\hrulefill} & \multicolumn{2}{c}{\hrulefill} & \multicolumn{2}{c}{\hrulefill} \\
& ACC (\%) & F1 (\%) & ACC Diff (\%) & F1 Diff (\%) & ACC Diff (\%) & F1 Diff (\%) & ACC Diff (\%) & F1 Diff (\%) & ACC Diff (\%) & F1 Diff (\%) \\
\midrule

\textbf{C6400\_N6\_M72} & \textbf{91.37} & \textbf{91.33} & \cellcolor{green!15} \textbf{86.08 [−5.29]} & \cellcolor{green!15} \textbf{86.33 [−5.00]} & \cellcolor{green!15} \textbf{76.49 [−14.88]} & \cellcolor{green!15} \textbf{76.01 [−15.32]} & \cellcolor{green!15} \textbf{86.09 [−5.28]} & \cellcolor{green!15} \textbf{86.24 [−5.09]} & \cellcolor{green!15} \textbf{82.42 [−8.95]} & \cellcolor{green!15} \textbf{82.58 [−8.75]} \\[2pt]
C5120\_N6\_M60 & 90.57 & 90.65 & \cellcolor{green!15} 83.48 [−7.09] & \cellcolor{green!15} 83.78 [−6.87] & \cellcolor{green!15} 76.85 [−13.72] & \cellcolor{green!15} 77.85 [−12.80] & \cellcolor{green!15} 83.69 [−6.88] & \cellcolor{green!15} 84.03 [−6.62] & \cellcolor{green!15} 76.86 [−13.71] & \cellcolor{green!15} 77.03 [−13.62] \\[2pt]
C6144\_N7\_M75 & 90.30 & 90.16 & \cellcolor{green!15} 84.24 [−6.06] & \cellcolor{green!15} 84.26 [−5.90] & \cellcolor{green!15} 75.91 [−14.39] & \cellcolor{green!15} 74.53 [−15.63] & \cellcolor{green!15} 86.82 [−3.48] & \cellcolor{green!15} 86.83 [−3.33] & \cellcolor{green!15} 83.03 [−7.27] & \cellcolor{green!15} 83.03 [−7.13] \\[2pt]
C5120\_N11\_M80 & 86.90 & 87.17 & \cellcolor{green!15} 68.62 [−18.28] & \cellcolor{green!15} 68.75 [−18.42] & \cellcolor{green!15} 64.83 [−22.07] & \cellcolor{green!15} 67.20 [−19.97] & \cellcolor{green!15} 66.21 [−20.69] & \cellcolor{green!15} 66.56 [−20.61] & \cellcolor{green!15} 69.31 [−17.59] & \cellcolor{green!15} 69.83 [−17.34] \\[2pt]
C7680\_N6\_M80 & 86.49 & 86.17 & \cellcolor{green!15} 71.77 [−14.72] & \cellcolor{green!15} 72.56 [−13.61] & \cellcolor{green!15} 67.73 [−18.76] & \cellcolor{green!15} 66.47 [−19.70] & \cellcolor{green!15} 72.98 [−13.51] & \cellcolor{green!15} 72.54 [−13.63] & \cellcolor{green!15} 65.71 [−20.78] & \cellcolor{green!15} 64.52 [−21.65] \\[2pt]
C6400\_N5\_M65 & 85.30 & 85.44 & \cellcolor{green!15} 80.24 [−5.06] & \cellcolor{green!15} 79.72 [−5.72] & \cellcolor{green!15} 70.74 [−14.56] & \cellcolor{green!15} 69.84 [−15.60] & \cellcolor{green!15} 73.91 [−11.39] & \cellcolor{green!15} 73.77 [−11.67] & \cellcolor{green!15} 76.73 [−8.57] & \cellcolor{green!15} 76.34 [−9.10] \\
\midrule
\multicolumn{11}{c}{\textit{... Additional 10 configurations ...}} \\
\midrule
C6144\_N6\_M70 & 77.98 & 78.59 & \cellcolor{green!15} 74.99 [−2.99] & \cellcolor{green!15} 75.16 [−3.43] & \cellcolor{green!15} 67.78 [−10.20] & \cellcolor{green!15} 70.61 [−7.98] & \cellcolor{green!15} 71.82 [−6.16] & \cellcolor{green!15} 73.70 [−4.89] & \cellcolor{green!15} 72.88 [−5.10] & \cellcolor{green!15} 74.10 [−4.49] \\[2pt]
\textbf{C10240\_N2\_M50} & \textbf{79.02} & \textbf{78.54} & \textbf{N/A} & \textbf{N/A} & \cellcolor{red!15} \textbf{81.53 [+2.51]} & \cellcolor{red!15} \textbf{81.27 [+2.73]} & \cellcolor{red!15} \textbf{88.85 [+9.83]} & \cellcolor{red!15} \textbf{88.86 [+10.32]} & \cellcolor{red!15} \textbf{79.31 [+0.29]} & \cellcolor{red!15} \textbf{78.83 [+0.29]} \\[2pt]
C5120\_N9\_M75 & 78.11 & 77.91 & \cellcolor{red!15} 83.00 [+4.89] & \cellcolor{red!15} 81.86 [+3.95] & \cellcolor{green!15} 73.76 [−4.35] & \cellcolor{green!15} 73.73 [−4.18] & \cellcolor{green!15} 77.14 [−0.97] & \cellcolor{green!15} 77.53 [−0.38] & \cellcolor{green!15} 68.20 [−9.91] & \cellcolor{green!15} 68.56 [−9.35] \\
\midrule
\multicolumn{11}{c}{\textit{... Additional 38 configurations ...}} \\
\midrule
C9600\_N13\_M95 & 55.86 & 55.50 & \cellcolor{red!15} 56.55 [+0.69] & \cellcolor{red!15} 56.72 [+1.22] & \cellcolor{green!15} 51.03 [−4.83] & \cellcolor{red!15} 60.74 [+5.24] & \cellcolor{red!15} 56.55 [+0.69] & \cellcolor{red!15} 59.83 [+4.33] & \cellcolor{red!15} 56.90 [+1.04] & \cellcolor{red!15} 59.30 [+3.80] \\[2pt]
C6144\_N151\_M99 & 53.64 & 54.98 & \cellcolor{red!15} 55.91 [+2.27] & \cellcolor{red!15} 55.34 [+0.36] & \cellcolor{green!15} 48.18 [−5.46] & \cellcolor{green!15} 49.77 [−5.21] & \cellcolor{red!15} 61.82 [+8.18] & \cellcolor{red!15} 61.21 [+6.23] & \cellcolor{red!15} 57.27 [+3.63] & \cellcolor{red!15} 56.56 [+1.58] \\[2pt]
C12800\_N21\_M99 & 55.88 & 54.73 & \cellcolor{green!15} 55.55 [−0.33] & \cellcolor{green!15} 52.52 [−2.21] & \cellcolor{green!15} 55.59 [−0.29] & \cellcolor{red!15} 58.02 [+3.29] & \cellcolor{red!15} 57.12 [+1.24] & \cellcolor{green!15} 53.07 [−1.66] & \cellcolor{green!15} 55.55 [−0.33] & \cellcolor{red!15} 56.32 [+1.59] \\[2pt]
C9600\_N61\_M99 & 54.30 & 52.71 & \cellcolor{red!15} 55.08 [+0.78] & \cellcolor{red!15} 57.57 [+4.86] & \cellcolor{green!15} 50.73 [−3.57] & \cellcolor{green!15} 49.47 [−3.24] & \cellcolor{red!15} 55.84 [+1.54] & \cellcolor{red!15} 59.18 [+6.47] & \cellcolor{green!15} 53.27 [−1.03] & \cellcolor{green!15} 52.62 [−0.09] \\
\bottomrule
\end{tabular}}
\smallskip
\begin{flushleft}
\small \textbf{Note:} Green cells (\raisebox{0.5ex}{\colorbox{green!15}{\hspace{1.5em}}}) indicate a decrease in performance relative to the RNN baseline for the same configuration; red cells (\raisebox{0.5ex}{\colorbox{red!15}{\hspace{1.5em}}}) indicate a performance increase. The values in square brackets represent the absolute difference from the RNN baseline for each configuration. N/A denotes configurations where evaluation for the corresponding model was not conducted.
\end{flushleft}

\label{tab:model-performance}
\end{table*}

To identify the best configuration for each deep learning model, we conducted a 5-fold cross-validation. As shown in Table~\ref{tab:cs_model_comparison}, two configurations emerged as top performers: C6400\_N6\_M72 with \gls{rnn}, and C10240\_N2\_M50 with \gls{lstm}. We evaluated \gls{rnn}, \gls{lstm}, \gls{gru}, \gls{cnn}-\gls{lstm}, and \gls{ae}-\gls{lstm} models, where simpler architectures such as the \gls{rnn} consistently outperformed more complex ones. Based on cross-validation results, we selected C6400\_N6\_M72 and C10240\_N2\_M50 for a second tuning phase focused on optimizing \gls{rnn}, \gls{lstm}, and \gls{gru} hyperparameters. As reported in Table~\ref{tab:best-configurations}, the \gls{rnn} with C10240\_N2\_M50 achieved the highest accuracy (95.60\%) and F1 score (95.69\%), followed by the same configuration with \gls{gru} (93.60\% / 93.80\%). C6400\_N6\_M72 also yielded strong results with \gls{rnn} (92.41\% / 92.00\%). Despite its ability to model long-term patterns, \gls{lstm} remained less effective and this shows the importance of the proposed model–configuration benchmark.

Overall, the results demonstrate that simpler models such as \gls{rnn} are more effective in both accuracy and computational efficiency, in this dataset. Consequently, we prioritized \gls{rnn}-based configurations for further analysis and deployment. The confusion matrix in Table~\ref{tab:confusion_matrix} illustrates the final test results for the selected \gls{rnn}-C6400\_N6\_M72 model.

\begin{table}[th]
\centering
\caption{Confusion matrix for the selected Simple\gls{rnn} model using configuration C6400\_N6\_M72. Model performance: Accuracy = 92\%, Precision = 97\%, Recall = 87\%, F1-score = 92\%.}
\label{tab:confusion_matrix}
\begin{tabular}{@{}lcc@{}}
\toprule
\multirow{2}{*}{\textbf{True Class}} & \multicolumn{2}{c}{\textbf{Predicted Class}} \\
\cmidrule(l){2-3}
 & \textbf{Negative (0)} & \textbf{Positive (1)} \\
\midrule
\textbf{Negative (0)} & \cellcolor{gray!20}77 & 2 \\
\textbf{Positive (1)} & 10 & \cellcolor{gray!20}69 \\
\bottomrule
\end{tabular}
\end{table}

\subsection{Performance of the physiology-enhanced adaptive braking strategy system}

By integrating drowsiness into the state space, the \gls{dddqn} agent effectively adapted to drowsy driving conditions, learning a braking strategy that responds appropriately across varying scenarios.

\begin{figure*}[t]
    \centering
    \includegraphics[width=0.49\linewidth]{"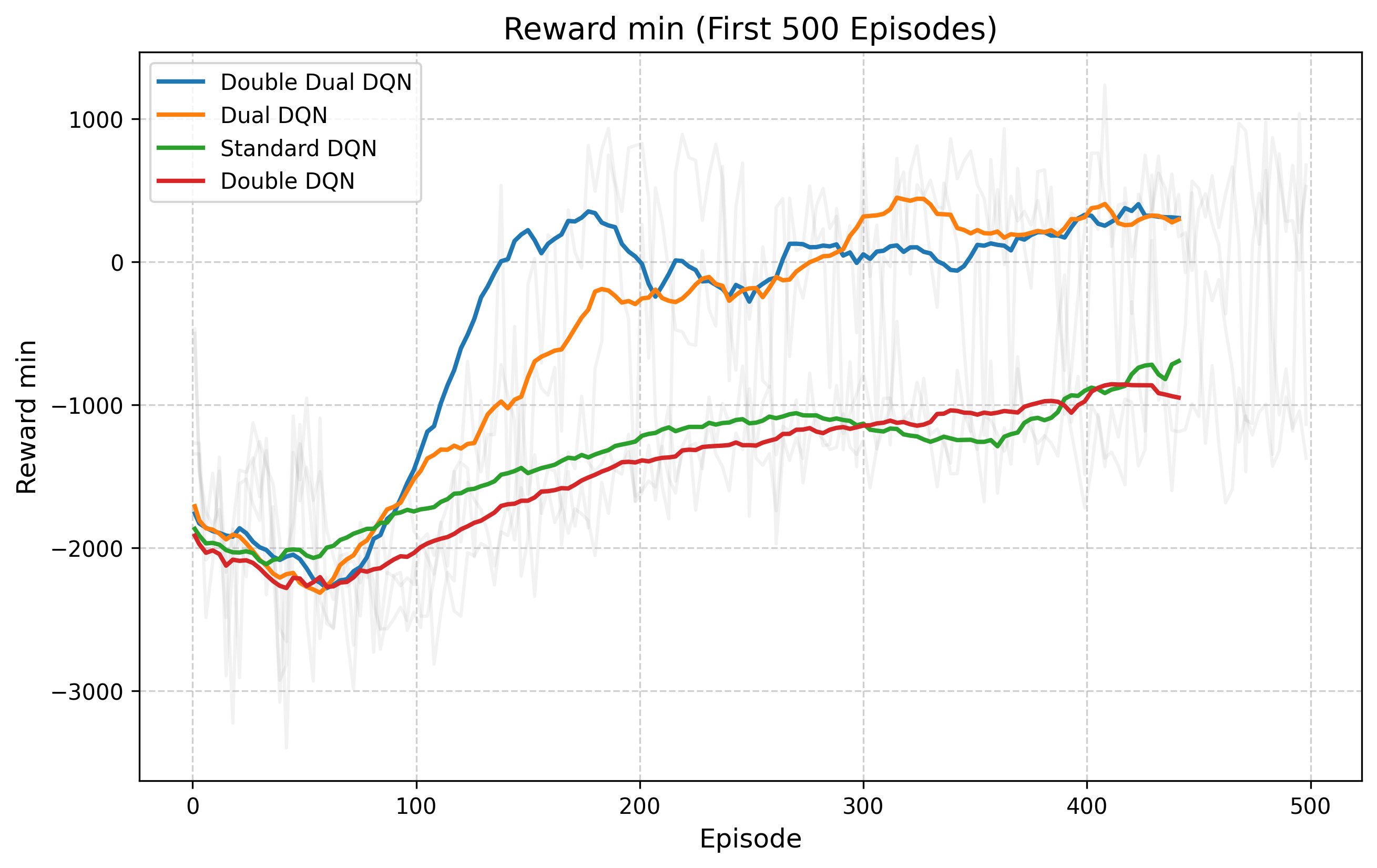"} 
    \includegraphics[width=0.49\linewidth]{"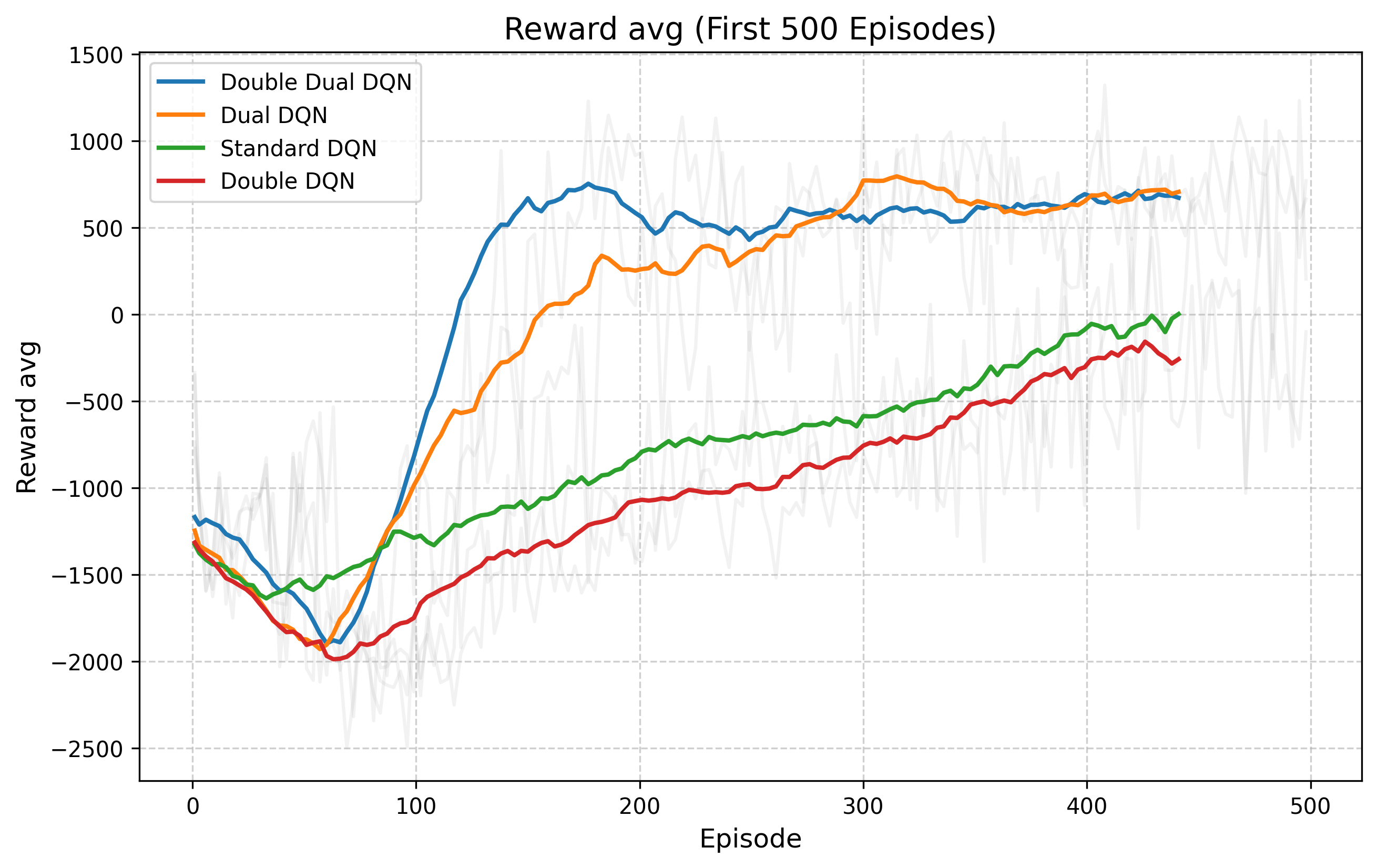"}
    \caption{Comparison of minimum and average rewards across training episodes.}
    \label{fig:reward-comparison}
\end{figure*}

The implemented \gls{dudqn} network uses three fully connected layers (\(128\), \(256\), and \(128\) units) to extract a shared representation, followed by parallel value and advantage streams, each with two layers (\(128\) and \(64\) units). Their outputs are combined using the standard \gls{dudqn} aggregation described in Section~\ref{sec:Background and Related Work}. To assess learning effectiveness, we first compare the training results of our \gls{dddqn} agent against the standard \gls{dqn} agent, \gls{dodqn} agent, and \gls{dudqn} agent. The plot in Figure~\ref{fig:reward-comparison} illustrates the average and minimum rewards over the first 500 training episodes, as this range is sufficient to observe differences in performance among the agents. A simple moving average is applied, where each cumulative rewards of episode data point is computed as the mean of the current and nine preceding values. This method reduces short-term fluctuations and facilitates the identification of overall trends. The minimum and average reward results highlight the effectiveness of the \gls{dddqn} framework reaching approximately 1500 and 500, respectively, by the 500th episode. For \gls{dodqn} and standard \gls{dqn}, both agents underperform compared to the \gls{dddqn} framework, with minimum and average rewards stabilizing at lower values around -500 and 200, respectively. This indicates that while \gls{dodqn} is designed to mitigate overestimation bias, it fails to leverage this advantage effectively in the given environment, resulting in a slower learning rate and suboptimal policy convergence. However, the combination of \gls{dodqn} with \gls{dudqn} in the \gls{dddqn} architecture addresses these shortcomings, as evidenced by its superior stability and rapid improvement in rewards during episodes from 100 to 200. The \gls{dddqn} agent not only outperforms \gls{dodqn} but also surpasses all other baselines, including \gls{dudqn}, which reaches an average reward plateau of around 700. The steep upward trend in \gls{dddqn}’s minimum rewards further reinforces its robustness, indicating fewer instances of poor decision-making and a reduced risk of policy degradation. These findings suggest that while \gls{dodqn} alone struggles to improve over \gls{dqn}, its integration within the \gls{dddqn} framework successfully enhances learning efficiency, leading to faster convergence and more reliable performance in complex environments.

To evaluate the agent's ability to recognize and adapt to drowsiness effects while maintaining safety, we tested the trained \gls{dddqn} agent on 1000 randomly generated scenarios. These were sampled from a larger space of possible combinations, considering variations in throttle pressure and inter-vehicle distance. The experiment also compared the agent's behavior under drowsy and non-drowsy conditions by analyzing matched pairs of episodes (i.e., the same scenarios executed once during drowsiness and once without it) to assess the impact of drowsiness on decision-making. The agent demonstrated a remarkably low failure rate of 0.1\%, with only one out of 1000 episodes failing to reach the 30-second brake test threshold, which aligns with the 30-second threshold described in Section~\ref{subsec:experimental_setup}. The 99.9\% success rate shows the agent effectively avoided collisions, both in the presence and absence of induced drowsiness effects. Additionally, the agent achieved an average cumulative reward of 1220 and a median cumulative reward of 1217, values that reflect consistent success across most scenarios. These numbers align with the reward function, which assigns high positive values only within the adjusted safe stopping distance, confirming the \gls{dddqn} agent’s adherence to the learned braking strategy.

\begin{figure}[t]
\centering
\includegraphics[width=\linewidth]{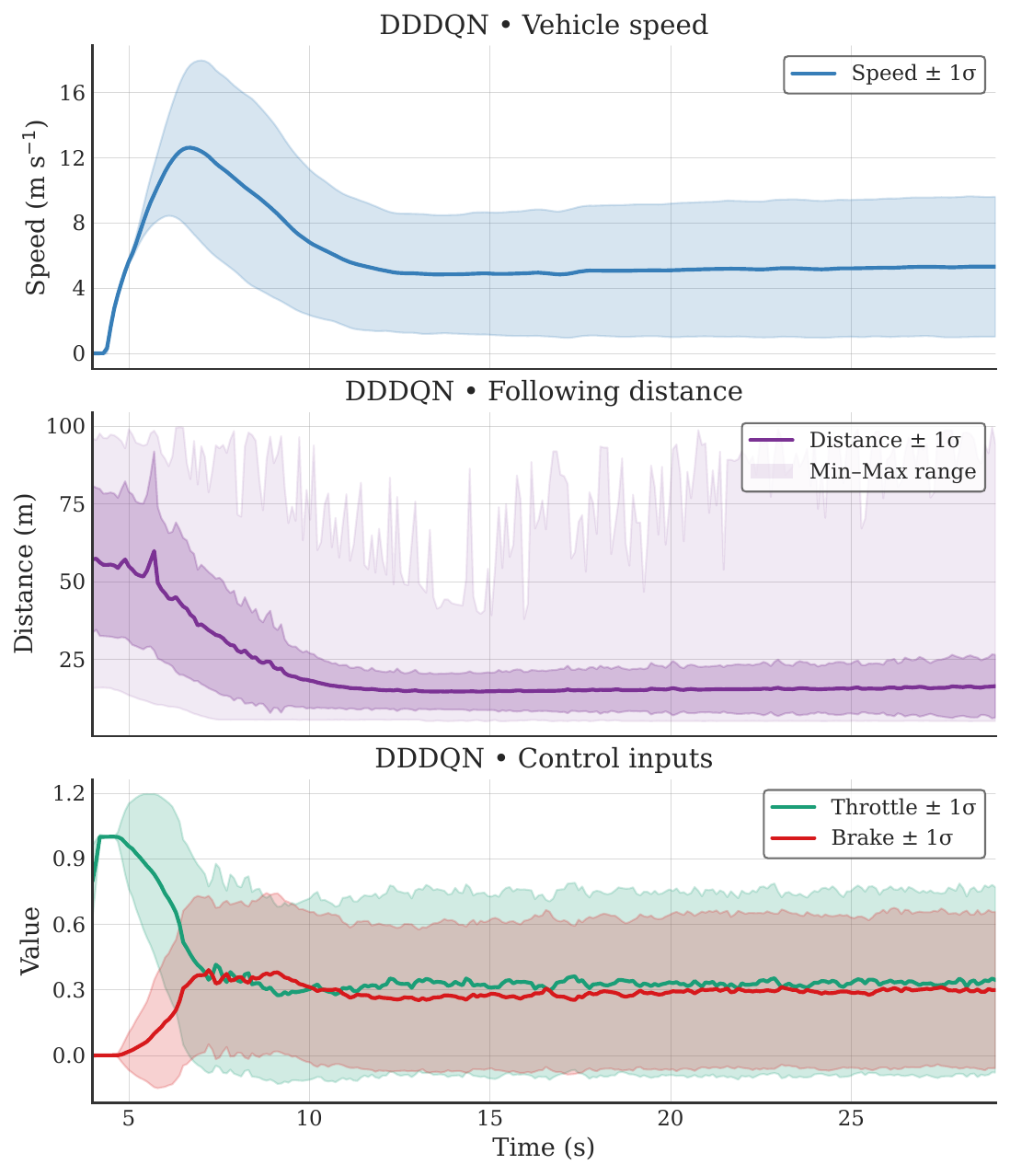}
\caption{Analysis of the DDDQN agent's behavior during 999 episodes including drowsy and non-drowsy scenarios. The figure presents the average ego vehicle speed with standard deviation (top), the relative following distance with min-max range (middle), and the control inputs (throttle and brake) with standard deviation (bottom). The plot highlights the agent's adaptive behavior across varied driving conditions.}
\label{fig:vehicle-dynamics-analysis}
\end{figure}

The general behavior of the agent across successful episodes is shown in Fig.~\ref{fig:vehicle-dynamics-analysis}, reflecting consistent performance with stable speed, safe following distance, and balanced control inputs under both drowsy and non-drowsy states. The plot highlights three key aspects: average vehicle speed with standard deviation (top), relative following distance with min-max range (middle), and control inputs (throttle and brake) with the standard deviation (bottom). Speed consistency appears as a stable velocity profile throughout the episodes, with minimal variation between tests. The following distance plot shows steady adherence to safe spacing, keeping the relative distance within the expected range across drowsy and non-drowsy conditions. Control inputs reveal a balanced use of throttle and brake, indicating the agent’s adaptive response to changing driving contexts. These visualizations demonstrate that the \gls{dddqn} agent's learned strategy remains robust across diverse scenarios.

The \gls{dddqn} agent successfully adapts to drowsy conditions, despite the 0.5 s action delay. Based on the three phases acceleration, braking, and following described in Section~\ref{subsec:dddqn_drowsiness}, the acceleration phase lasts 24.6\% longer for drowsy episodes, yet the agent maintains safe control with no collisions even when actions are delayed. The \gls{dddqn} agent kept an average safe distance of 22.89 m, nearly identical to the 22.90 m observed in non-drowsy cases, based on the average across all tested scenarios. In braking strategy, the agent compensates with a slightly higher average braking pressure (0.28 vs. 0.25) to correct deviations caused by impaired decision-making.

\begin{figure*}[th]
    \centering
    \includegraphics[width=\linewidth]{"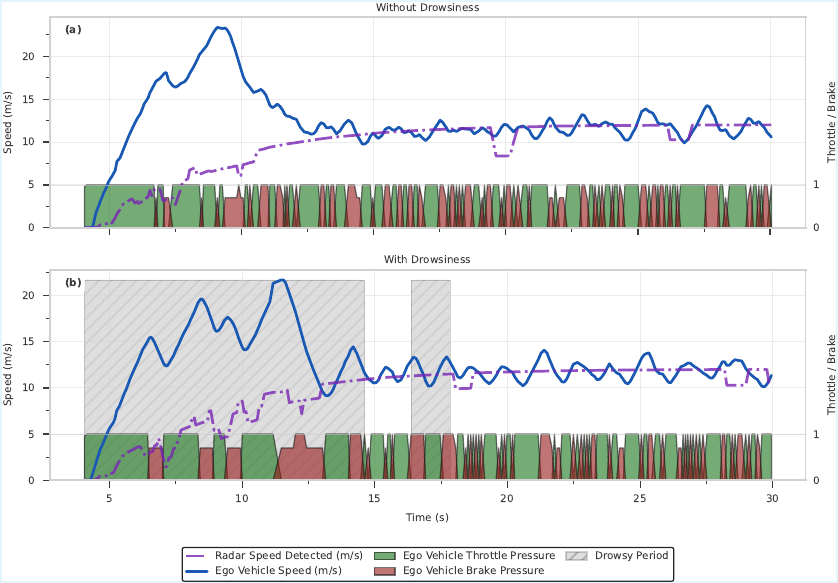"}
    \caption{Impact of drowsiness on the control behavior of the ego vehicle driven by the agent. (a) illustrates the agent's control strategy under normal conditions, while (b) highlights the effect of drowsiness, showing drowsy periods (shaded regions) and variations in the ego vehicle’s throttle and brake inputs in response to the front vehicle’s speed.
}
    \label{fig:drowsiness_impact}
\end{figure*}
Figure~\ref{fig:drowsiness_impact} compares the agent's adaptive control strategy in two paired scenarios where (a) represents the baseline without drowsiness and (b) incorporates drowsiness-induced action delays. This scenario pair highlights the robustness of the learned policy in mitigating cognitive impairment effects on vehicle control dynamics while maintaining longitudinal stability and safe car-following behavior. Despite the delay perturbation in (b), the agent effectively regulates its following distance by adopting a low-variability control strategy, prioritizing smoother throttle and braking adjustments to ensure stability. However, this adaptive response leads to greater difficulty in speed alignment with the front vehicle, as delayed actions reduce acceleration and deceleration responsiveness. These learned adjustments enable the agent to maintain a safe braking zone while minimizing the risk of instability. Although delayed responses impact reaction times, the policy compensates by dynamically adjusting braking intensities to mitigate deviations from the optimal car-following distance. 

To further quantify the effect of drowsiness on stability, we computed the total time spent by the agent in violated unsafe distance. Out of 30,000 seconds of testing, only 0.9 seconds of cumulative breaches occurred, exclusively under drowsy conditions. The results underscore the agent’s \gls{dddqn} framework in accommodating human-like cognitive delays, demonstrating its ability to preserve safety and efficiency in realistic driving scenarios affected by drowsiness-induced delays. These findings reinforce the potential of intelligent driver-assist systems in enhancing road safety under cognitive impairment conditions.

\section{Discussion and Conclusion}
\label{sec:Discussion}

Drowsy driving increases the risk of collisions due to delayed reactions. We propose an adaptive braking system that integrates drowsiness-aware signals into a \gls{dddqn} agent. The agent learns effective braking strategies despite delayed actions, using \gls{ecg}-based detection and discrete braking levels for closed-loop control. 

The results from the drowsiness detection module reveal that temporal models—particularly the \gls{rnn}—are capable of achieving high discriminative performance with minimal architectural complexity. Our proposed optimal capsule-shifting approach reliably contributes to distinguish drowsy events, which supports the hypothesis that fine-grained short-term temporal dependencies in \gls{hrv} signals are critical to detecting cognitive fluctuations. The comparative degradation observed in more complex models like \gls{cnn}-\gls{lstm} or \gls{ae}-\gls{lstm} across most configurations suggests that these architectures may suffer from overfitting in the presence of limited training variability or fail to capture short-term transient dynamics essential for drowsiness detection. The \gls{dddqn} agent outperformed \gls{dqn}, \gls{dodqn}, and \gls{dudqn} baselines, demonstrating robust collision avoidance and adaptive behavior. By embedding physiological signals, the agent compensated for delayed responses through smoother transitions and braking during drowsy episodes. Despite longer acceleration phases during drowsiness delays, safe distances were maintained by the agent. Episode rewards confirmed performance stability, validating the benefit of integrating cognitive delay directly into learning.

However, several critical limitations constrain the scalability and external validity of this approach. On the physiological side, the drowsiness model is trained on a relatively constrained ECG dataset using a fixed two-minute event window, which may not capture subtle or short-lived transitions in cognitive state. Consequently, the system may underperform in real-world deployments where driver behavior evolves dynamically and is influenced by multiple, overlapping sources of fatigue, distraction, or stress. Furthermore, the capsule-shifting strategy, while effective for event labeling, depends heavily on the structure and density of labeled events, which may limit its sensitivity under different sampling conditions. From an agent-centric perspective, a single lead vehicle and fixed-speed profiles limit the agent’s exposure to real-world traffic complexity. While this abstraction facilitates interpretability and accelerates training, it limits the model’s exposure to more nuanced driving interactions, such as sudden cut-ins, varying lane dynamics, or social negotiation behaviors typical in urban and multi-agent scenarios.

Looking forward, two promising directions emerge. First, the physiological model can be improved by incorporating a broader set of biosignals, dynamic windowing techniques, and larger datasets that capture a wider spectrum of driver states and reactions. This would enable a more continuous and fine-grained representation of cognitive impairment, feeding richer context into the control agent. Second, expanding the simulation to support decentralized multi-agent settings with vehicle-to-vehicle communication and partial observability could better approximate real-world constraints. In such settings, coordination becomes critical, and the agent would need to balance self-preservation with cooperative safety, particularly under asymmetric drowsiness across agents.

In conclusion, this work presents a principled and effective framework for physiology-aware braking control using deep reinforcement learning. By integrating drowsiness recognition with an adaptive control policy, we demonstrate the potential to mitigate the risks of cognitive delay in longitudinal vehicle control. The observed results support the broader vision of embedding human-centered feedback loops within autonomous decision-making, paving the way for safer, more personalized, and cognitively adaptive intelligent driving systems.

\section{Acknowledgments}
This work has been partially funded by the “MUR, Ministero dell’Universita e della Ricerca” Ministerial Decree n. 737 dated 25-06-2021 “Criteri di riparto e utilizzazione del Fondo per la promozione e lo sviluppo delle politiche del Programma Nazionale per la Ricerca (PNR)” under the project title: “IDAS – Innovazione Digitale in Ambito Salute”- CUP:
F84D22000270001.
In addition, it has been partially funded by the grant RYC2021-032853-I from MCIN/AEI/ 10.13039/501100011033, the European Union NextGenerationEU/PRTR, and the European Union’s Horizon Europe research and innovation programme under the Marie Sklodowska-Curie actions (Grant Agreement 101202439 - SWEATHEART). Finally, authors would like to thank IIT (Istituto Italiano
di Tecnologia) for its funding support of Hossem Eddine Hafidi.

\nocite{*}
\bibliographystyle{IEEEtran}
\bibliography{references}

\begin{IEEEbiography}[{\includegraphics[width=1in,height=1.25in,clip,keepaspectratio]{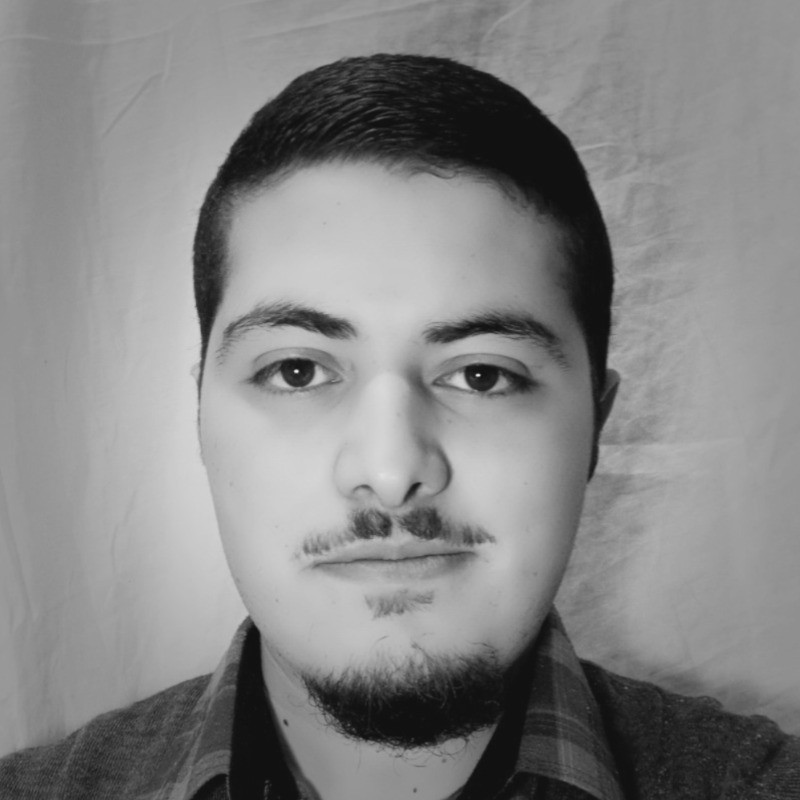}}]{Hossem Eddine Hafidi}

is a PhD student jointly affiliated with the Istituto Italiano di Tecnologia (IIT) and the IDentification Automation Laboratory (IDA Lab), Department of Innovation Engineering, University of Salento, Italy. He earned his BSc in Computer Science in 2019 and his MSc in Software Engineering and Distributed Systems in 2021 from the University of Mohamed Khider Biskra, with a major in Software Engineering. His current research focuses on the integration and correlation of vehicular dynamics, physiological signals, and environmental context through IoT and Edge AI to support real-time healthcare monitoring in the automotive domain.

\end{IEEEbiography}

\begin{IEEEbiography}[{\includegraphics[width=1in,height=1.25in,clip,keepaspectratio]{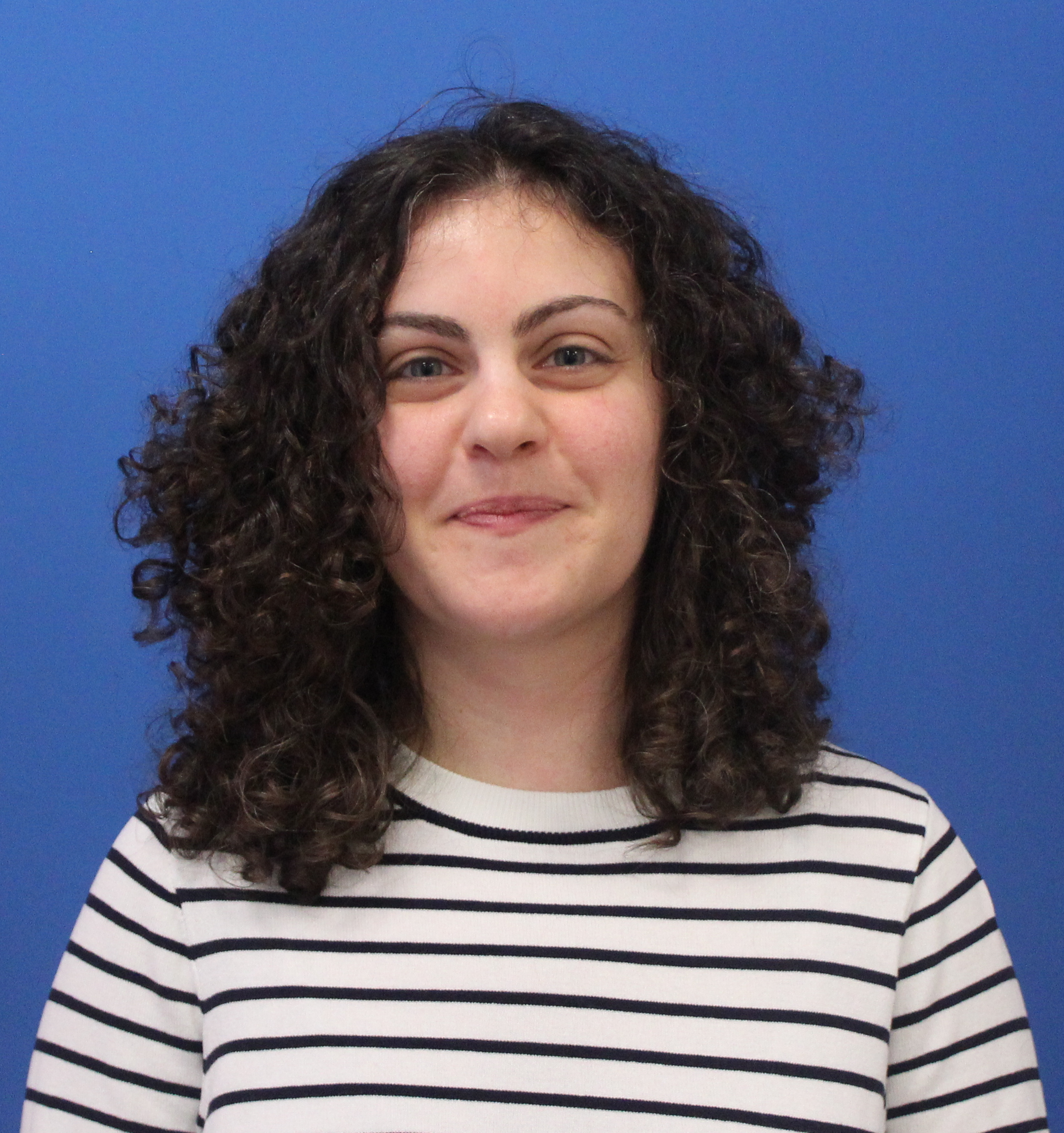}}]{Elisabetta De Giovanni}
is an MSCA Postdoctoral Fellow at the Basque Center for Applied Mathematics (BCAM) in Bilbao, Spain, focusing on energy-efficient wearable systems to monitor cardiorespiratory effects in elite athletes. She holds a master's degree in Bioengineering from the University of Pavia and earned her PhD in Electrical Engineering from École Polytechnique Fédérale de Lausanne (EPFL) in 2021, specializing in adaptive ultra-low power wearable sensors for health monitoring. After a year-long career break in 2022, she worked as a postdoctoral fellow at the University of Salento, developing intelligent IoT solutions for health, sport, and safety. Elisabetta has authored multiple articles in top journals and conferences, received over 100 citations, peer-reviewed various articles for international journals, and contributed to numerous research projects.
\end{IEEEbiography}

\begin{IEEEbiography}[{\includegraphics[width=1in,height=1.25in,clip,keepaspectratio]{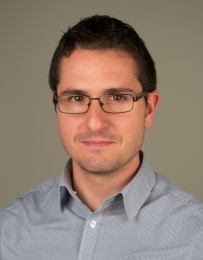}}]{Teodoro Montanaro} (Member, IEEE, since 2015) received the M.S. degree in Computer Engineering in 2014 and the Ph.D. degree in Computer and Control Engineering in 2018 from the Politecnico di Torino, Turin, Italy. From 2017 to 2020 he collaborated with the Istituto Superiore Mario Boella (ISMB) and LINKS Foundation by cooperating to different researches funded by the European Commission like IoF2020, S4G, MONICA, MAESTRI that brought innovations in different fields (e.g., food and farm, grid, city, health). Since 2020, he collaborates with the IDentification Automation Laboratory (IDA Lab), Department of Innovation Engineering, Università del Salento. His current research interests include IoT applications specifically focused on the exploitation of fog computing, DLT, blockchain, AI and Federated Learning in different domains like smart grids, smart homes, smart cities, industrial processes, food traceability, and smart health. He has authored different papers on international journals and conferences. He has co-chaired some tracks, workshops, and symposia within the IEEE COMPSAC, the IEEE WF-IoT and the IEEE Splitech conferences. He has also contributed to review some papers in some important Journals.
\end{IEEEbiography}

\begin{IEEEbiography}[{\includegraphics[width=1in,height=1.25in,clip,keepaspectratio]{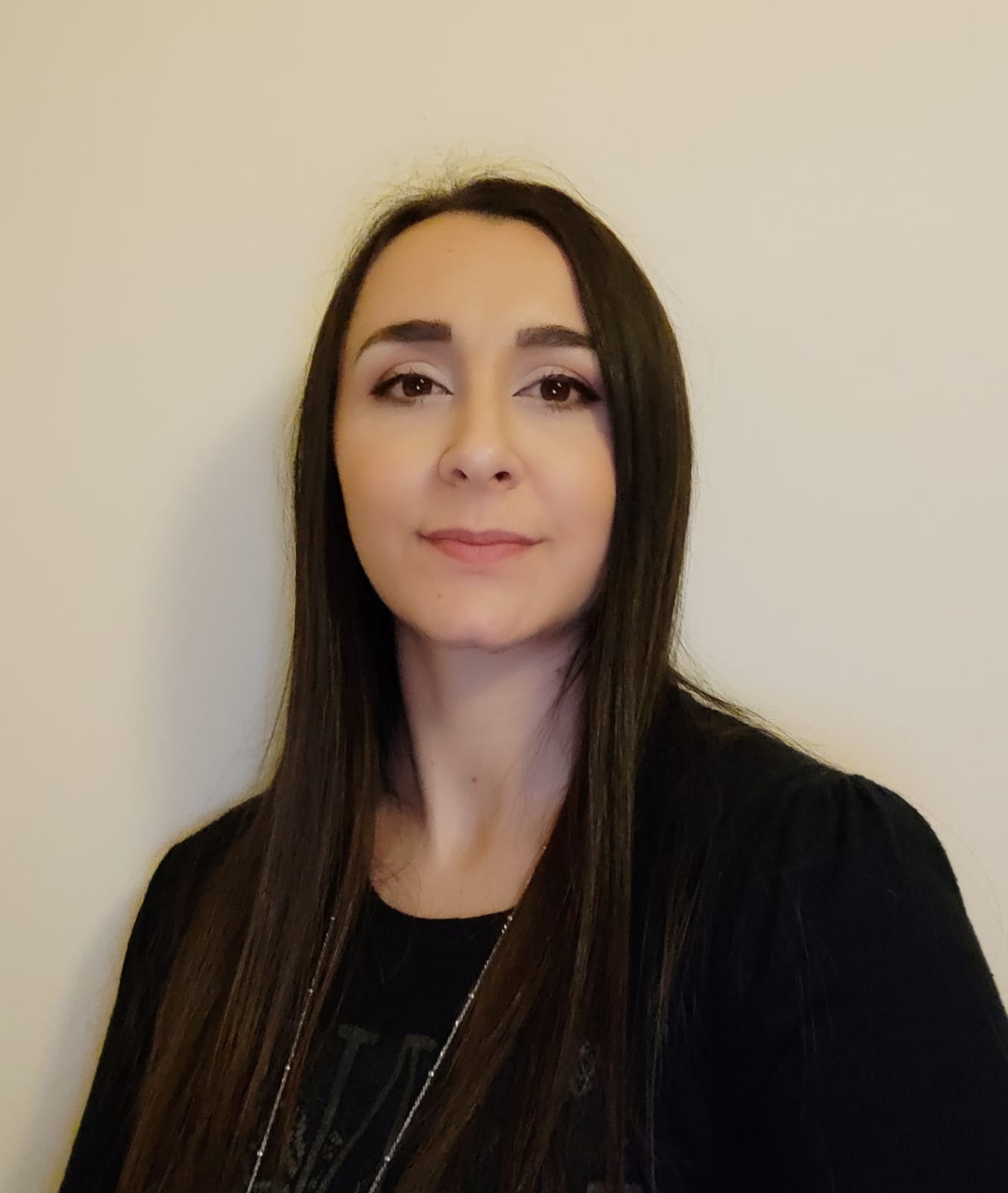}}]{Ilaria Sergi}
received the master’s degree in Automation Engineering from the University of Salento, Lecce, Italy, in 2012.  Her thesis focused on the tracking of small laboratory animals, based on passive UHF RFID technology. Since 2012, she collaborates with the Identification Automation Laboratory (IDA Lab) of the Department of Engineering for Innovation, University of Salento. In 2019 she received the PhD in Engineering of Complex Systems from the University of Salento, Italy. Her research interests include RFID, Bluetooth, Internet of Things, smart environments, and homecare solutions. She has authored several papers on international journals and conferences.
\end{IEEEbiography}

\begin{IEEEbiography}[{\includegraphics[width=1in,height=1.25in,clip,keepaspectratio]{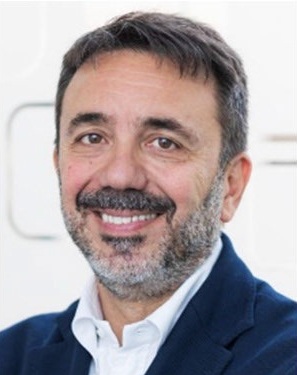}}]{Massimo de Vittorio}
(Senior Member, IEEE) is a Coordinator with the Center for Biomolecular Nanotechnologies, Italian Institute of Technology (IIT), Arnesano, Italy, and a Full Professor at Salento University, Lecce, Italy. He has authored 340 articles in indexed journals, 14 patents,
and 60 invited talks to international conferences. His research interests include activity deals with the development of nano-photonics and -electronics, micro-electro-mechanical systems, flexible sensors for healthcare, energy, and information and communication technologies (ICT). \end{IEEEbiography}

\begin{IEEEbiography}[{\includegraphics[width=1in,height=1.25in,clip,keepaspectratio]{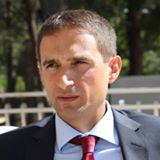}}]{Luigi Patrono}
(Member, IEEE) received the M.S. degree in computer engineering from the University of Lecce, Lecce, Italy, in 1999, and the Ph.D. degree in innovative materials and technologies for satellite networks from the ISUFI-University of Lecce, Italy, in 2003. He is an Associate Professor of Computer Networks and Internet of Things at the University of Salento, Lecce, where he is also the Pro-Vice Chancellor for Digital Technologies. His research interests include RFID, IoT, wireless sensor networks, and embedded systems. He has authored more than 160 scientific papers published in international journals and conferences. He has been the Organizing Chair of some international symposia and workshops, technically co-sponsored by the IEEE Communication Society, focused on the Internet of Things.
\end{IEEEbiography}

\end{document}